\documentclass[10pt,twocolumn,letterpaper]{article}

\usepackage[pagenumbers]{cvpr} 

\usepackage{graphicx}
\usepackage{amsmath}
\usepackage{amssymb}
\usepackage{booktabs}
\usepackage{xcolor} 
\usepackage{listings} 
\usepackage{enumitem} 
\usepackage{multirow} 
\usepackage{listings}  
\usepackage{xspace}  
\setlist[itemize]{noitemsep}
\usepackage{makecell}
\usepackage{todonotes}
\usepackage{pifont}
\newcommand{\ballotx}{\ding{55}}      
\newcommand{\ballotcheck}{\ding{51}}  

\xspaceaddexceptions{’}

\usepackage[pagebackref,breaklinks,colorlinks]{hyperref}

\usepackage[capitalize]{cleveref}
\crefname{section}{Sec.}{Secs.}
\Crefname{section}{Section}{Sections}
\Crefname{table}{Table}{Tables}
\crefname{table}{Tab.}{Tabs.}

\usepackage{xcolor}

\newcommand{\eclair}{\'{E}CLAIR\xspace}
\newcommand{\benchds}{DROBS\xspace}

\newcommand{\sred}[1]{\textcolor{red}{\texttt{#1}}}
\newcommand{\sblue}[1]{\textcolor{blue}{\texttt{#1}}}
\newcommand{\sgreen}[1]{\textcolor{ForestGreen}{\texttt{#1}}}
\newcommand{\spurple}[1]{\textcolor{purple}{\texttt{#1}}}

\begin{document}

\title{\eclair{} -- Extracting Content and Layout with Integrated \newline Reading Order for Documents}

\author{
    Ilia Karmanov\thanks{Equal contribution.}, Amala Sanjay Deshmukh{\footnotemark[1]}, Lukas Vögtle, Philipp Fischer, Kateryna Chumachenko, \\Timo Roman, Jarno Seppänen, Jupinder Parmar, Joseph Jennings,  Andrew Tao, Karan Sapra {\thanks{Project Lead.}}\\
    NVIDIA\\
    \texttt{\small \{ikarmanov,amalasanjayd,lvoegtle,pfischer,kchumachenko, }\\
    \texttt{\small troman,jseppanen,jupinderp,jjennings,atao,ksapra\}@nvidia.com}
}

\maketitle

\begin{abstract}

Optical Character Recognition (OCR) technology is widely used to extract text from images of documents, facilitating efficient digitization and data retrieval. However, merely extracting text is insufficient when dealing with complex documents. Fully comprehending such documents requires an understanding of their structure — including formatting, formulas, tables, and the reading order of multiple blocks and columns across multiple pages — as well as semantic information for detecting elements like footnotes and image captions. This comprehensive understanding is crucial for downstream tasks such as retrieval, document question answering, and data curation for training Large Language Models (LLMs) and Vision Language Models (VLMs). To address this, we introduce \eclair, a general-purpose text-extraction tool specifically designed to process a wide range of document types. Given an image, \eclair is able to extract formatted text in reading order, along with bounding boxes and their corresponding semantic classes. To thoroughly evaluate these novel capabilities, we introduce our diverse human-annotated benchmark \benchds for document-level OCR and semantic classification. \eclair achieves state-of-the-art accuracy on this benchmark, outperforming other methods across key metrics. Additionally, we evaluate \eclair on established benchmarks, demonstrating its versatility and strength across several evaluation standards.

\end{abstract}

\begin{figure*}[ht]
    
    \centering
    \begin{subfigure}{0.55\columnwidth}
        \centering
        \includegraphics[width=\linewidth]{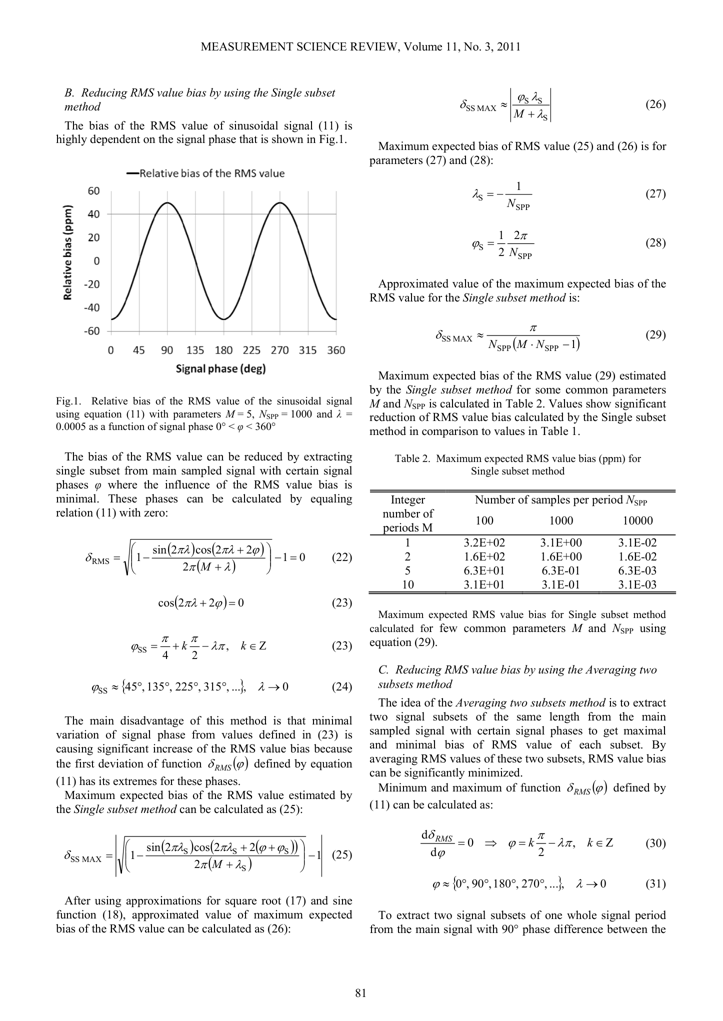} 
        \caption{}
        \label{fig:subfig_a}
    \end{subfigure}
    \hfill
    \begin{subfigure}{0.55\columnwidth}
        \centering
        \includegraphics[width=\linewidth]{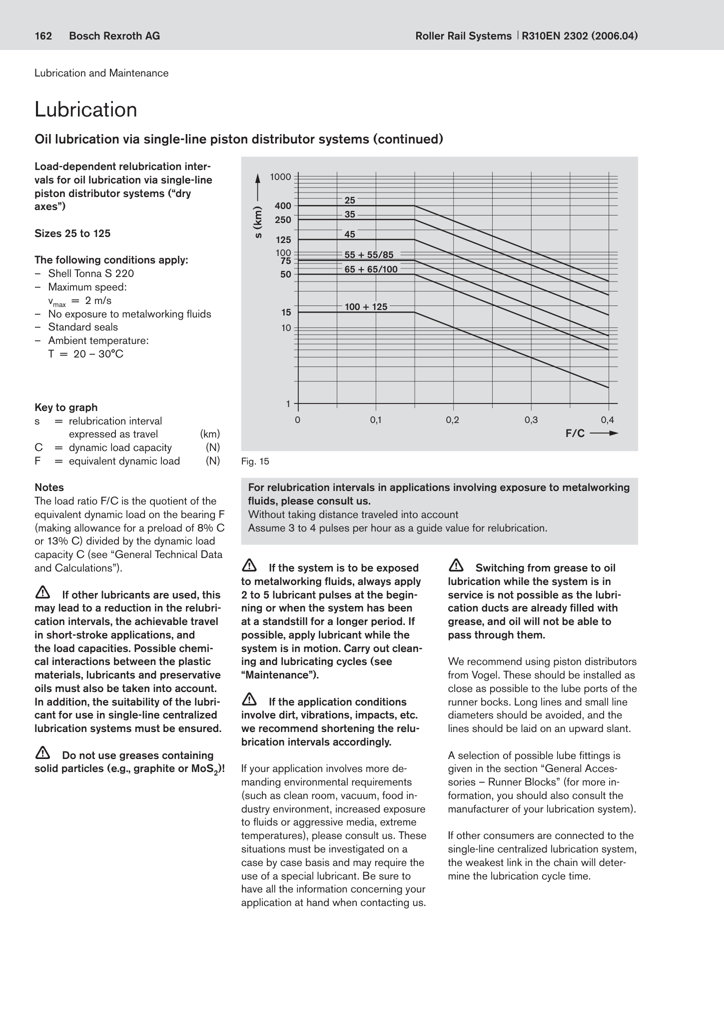} 
        \caption{}
        \label{fig:subfig_b}
    \end{subfigure}
    \hfill
    \begin{subfigure}{0.55\columnwidth}
        \centering        
        \includegraphics[width=\linewidth]{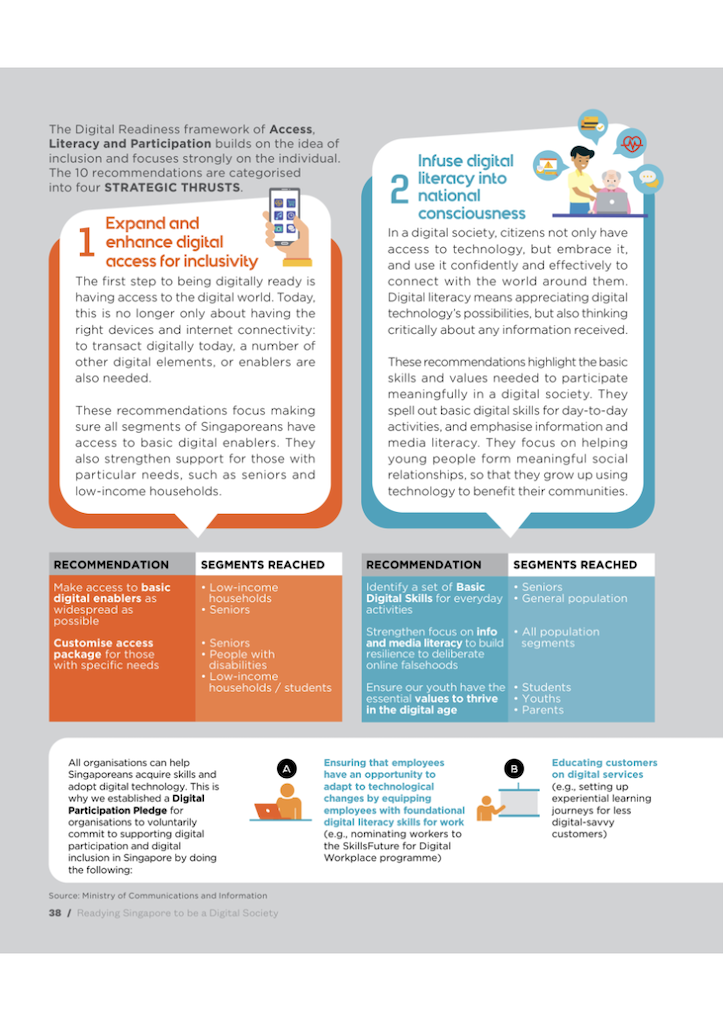} 
        \caption{}
        \label{fig:subfig_c}
    \end{subfigure}
    \caption{\eclair outperforms other methods on complex documents: (a)~tables, formulas, figure, page header and multiple columns; (b)~uneven columns, styling, figure; (c) non-obvious reading order and visual elements like background coloring.}
    \label{fig:teaser}
    \vspace{-5mm}
\end{figure*}
\vspace{-5mm}
\section{Introduction}

Optical Character Recognition (OCR) has allowed machines to extract text from images and transformed the way we interact with textual information. The recent success of Large Language Models (LLMs) is partly attributed to the availability of extremely large text datasets, placing an increasing demand for high-quality tokens extracted from text-dense documents such as scientific textbooks and journals. This is a challenging task since it necessitates an understanding of reading order across complex layouts, which in turn requires identifying different semantic ele\-ments and their relationships on the page. Maintaining a seamless flow requires separating relevant ele\-ments (e.g.,  paragraphs or tables) from irrelevant ones (e.g., page headers, page footers and other floating text).

Traditional OCR systems operate on a word or line level and are unable to properly understand the spatial and semantic relationships and hierarchies present in text-dense documents. More complex systems that possess such capabilities are generally not end-to-end and combine several models into a brittle pipeline. This shortcoming has spawned an interest in developing end-to-end models \cite{Kosmos, GOT, Nougat} that can extract formatted and structured text from complex documents such as those shown in Figure~\ref{fig:teaser}. Such capabilities provide downstream benefits for a multitude of tasks, including retrieval, document question answering, and increasing the availability of text tokens for LLM training. However, recent models proposed in this area still have crucial drawbacks: Kosmos-2.5~\cite{Kosmos} lacks the ability to extract formatted text that is at the same time spatially aware (since it has two mutually-exclusive prompts), while GOT~\cite{GOT} and Nougat~\cite{Nougat} do not predict any spatial information at all. In addition, none of these models predict semantic classes of bounding boxes, which can be used as a conditioning element for retrieval, help filter out irrelevant information for LLM training, and assist when combining multiple pages within a document (e.g., placing footnotes only after a text section has ended).

To address these concerns, we present \eclair: a multi-modal LLM (MLLM) comprised of a ViT-like encoder and an auto-regressive decoder, architecturally similar to Donut~\cite{donut}. \eclair is able to extract text (formatted as markdown/{\LaTeX}), bounding boxes of text blocks with their semantic classes, and any combination of these simultaneously, while preserving the reading order.

Training such a versatile model necessitates a dataset that encompasses all these annotation types. To address this problem, we generate arXiv-5M, a large-scale dataset that is sampled from arXiv papers, covers all desired annotation capabilities and serves as a link between existing datasets of varying annotation coverage.

Our proposed novel data generation pipeline includes a modified {\LaTeX} compiler which generates ground truth labels directly from the {\LaTeX} sources. 

Furthermore, existing document-level OCR benchmarks are limited by partial annotations: GOT~\cite{GOT} allows for document level reading order but lacks block-level spatial and semantic labels, while DocLayNet~\cite{Doclaynet} lacks reading order information. To address these shortcomings, we release a new benchmark \benchds consisting of 789 visually-diverse pages sampled from different sources, see Figure~\ref{fig:test-set-examples} for examples. The annotations come from human labeling and contain text in reading-order along with bounding boxes and semantic classes. \eclair achieves state-of-the-art (SOTA) accuracy on \benchds when compared to other recent models, as well as competitive metrics on several existing benchmarks spanning different tasks, including general OCR, document layout understanding, and extraction of high-quality data for LLM training.

To summarize, our contributions are as follows:
\begin{itemize}
    \item We create an end-to-end document-level OCR model which is the first to extract formatted text with their respective bounding boxes and semantic classes.
    \item We develop a novel data generation pipeline where we can control the rendering of {\LaTeX} sources ourselves. This is a pre-requisite for bridging the gap between existing datasets with fewer label types.
    \item We release a new benchmark \benchds, with high-quality human-labeled data and show SOTA accuracy as well as competitive metrics on existing benchmarks.
\end{itemize}

\begin{figure*}[t]
  \centering
    \begin{tabular}{@{} l l l @{}}
      a) & \begin{minipage}{0.83\textwidth}\includegraphics[width=\textwidth]{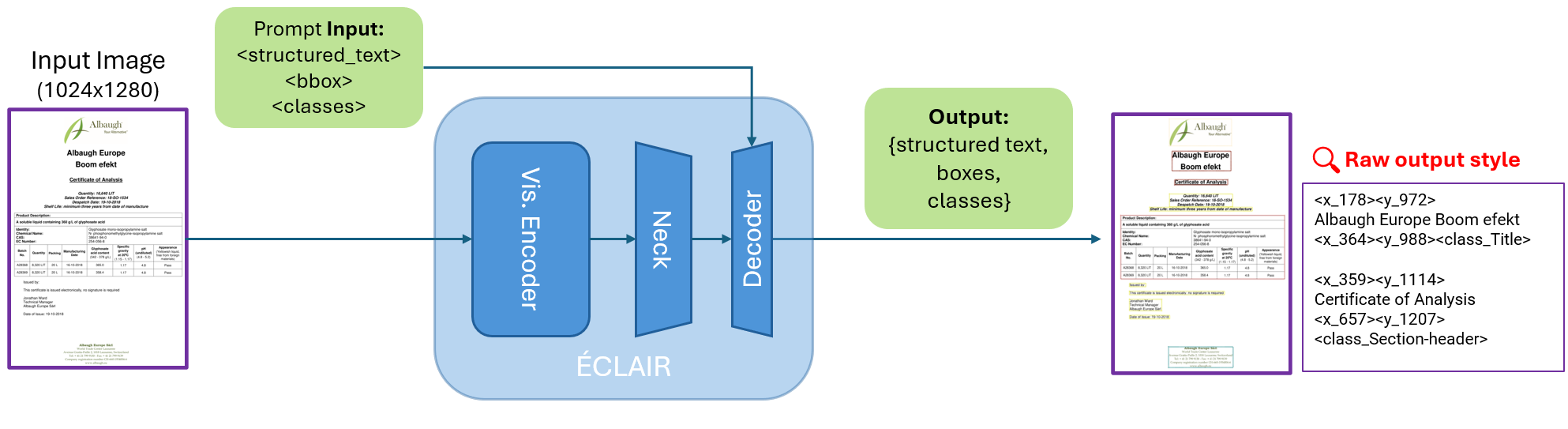}\\* \hrule\end{minipage} & \multirow{2}{*}{\includegraphics[width=0.1\textwidth]{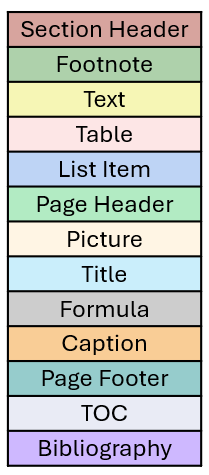}} \\
      b) & \begin{minipage}{0.83\textwidth}\vspace{0.5em}\includegraphics[width=\textwidth]{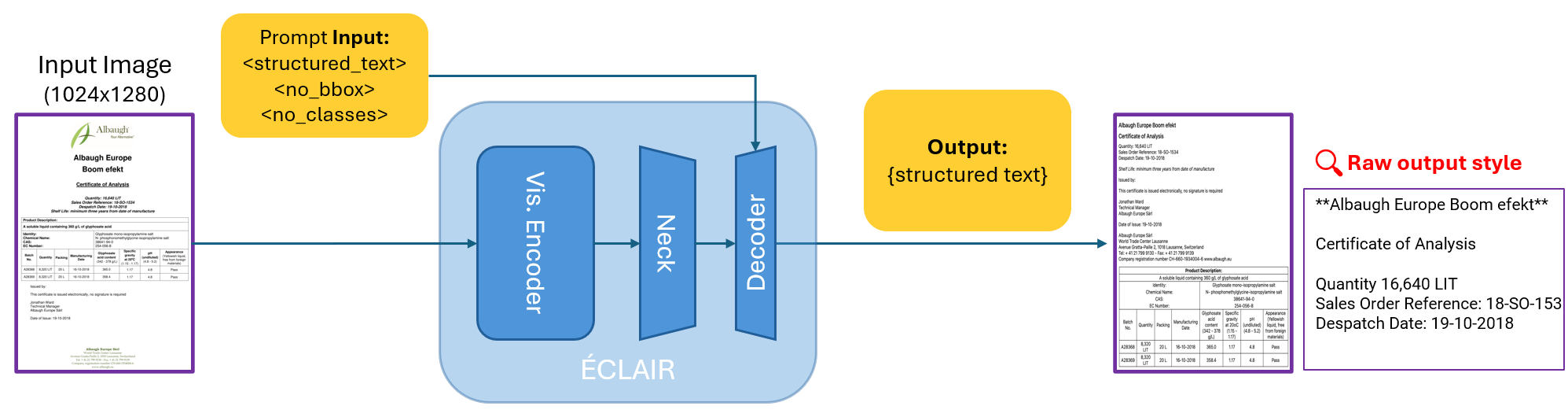}\end{minipage} & \\
    \end{tabular}
    \vspace{-2mm}
  \caption{Meta architecture for \eclair showcasing the usage with two different (out of eight valid) prompts: Example a) uses the maximal information prompt to return bounding boxes along with their semantic class, markdown text, and tables and formulas. In b) we ask the model to return only markdown text without boxes or classes. All supported semantic classes are listed on the right.}
  \label{fig:eclair_prompts}
  \vspace{-5mm}
\end{figure*}

\section{\eclair}

\subsection{Architecture}

\eclair uses a transformer encoder-decoder architecture. The \textbf{vision encoder}, denoted as $\mathcal{E}$, is initialized from RADIO~\cite{radio} which follows a ViT-H /16 \cite{vit} architecture, and maps an image  $\mathbf{I} \in \mathbb{R}^{3 \times H \times W}$ to a latent representation $\mathbf{Z} \in \mathbb{R}^{N \times d}$, where $d$ is the hidden dimension and $N$ is the sequence length. The \textbf{neck} $\mathcal{N}$ then reduces the dimensionality of the latent space as well as the sequence length.

The \textbf{decoder}, denoted as $\mathcal{D}$, uses mBART \cite{mbart} and predicts text tokens $\mathbf{T} = \{t_{P+1}, t_{P+2}, \dots, t_L\}$ by conditioning on the latent encoder representation, $\mathcal{N}(\mathbf{Z})$, and the context $t_{<i}$, $P(t_i | \mathcal{N}(\mathbf{Z}), t_{<i})$, where $\mathbf{Z} = \mathcal{E}(\mathbf{I})$ and $\{t_1, t_2, \dots, t_P\}$ are the prompt tokens and where $L$ is the prompt-augmented sequence length.

Since autoregressive models scale poorly with the decoder size and sequence length at inference time, we adopt a larger vision encoder (657M parameters) and combine it with a lightweight decoder (279M parameters). This follows from the observation that OCR is not fundamentally a generative task but rather depends on the content in the input image. We describe further modifications to improve inference time in Section~\ref{multi_token_inference}.

\subsection{Prompt}
\label{sec:prompt}

We use the input prompt to specify the desired format of the model outputs. Each prompt is a tuple of three options, with 8 possible valid combinations (ignoring the trivial case of no output, and the cases where semantic classes are requested without the corresponding bounding boxes):

\begin{flushleft}
\begin{small}
\begin{itemize}
    \item \verb|<structured_text>| or \verb|<plain_text>| or~\verb|<no_text>|
    \item \verb|<bbox>| or \verb|<no_bbox>|
    \item \verb|<classes>| or \verb|<no_classes>|
\end{itemize}
\end{small}
\end{flushleft}

For each of the three groups, the first option specifies the most information, while the last option suppresses this output type. With the \verb|<structured_text>| prompt, the text is predicted in markdown format and inline formulae are formatted as \LaTeX, whereas with the \verb|<plain_text>| prompt both are formatted as plain text. Tables and block formulae are formatted as {\LaTeX} for both modes. We define the maximal-information prompt (MIP) as: \begin{lstlisting}[basicstyle=\ttfamily,breaklines=true]
<structured_text><bbox><classes>
\end{lstlisting}

The novelty of \eclair compared to existing methods lies in its ability to handle any of the 8 valid prompt combinations. This is achieved by pre-training on a custom dataset that has labels for the maximal-information setting and then decreasing the information density for each group with some dataset-dependent probability during the fine-tuning stage. This allows the model to leverage visually diverse datasets with partial annotations for training. A schematic structure of \eclair along with possible prompts and corresponding output is presented in Figure \ref{fig:eclair_prompts}. 

\subsection{Output Format and Tokenization}
\label{output_format}

\eclair predicts bounding boxes of the semantic blocks in the form of discrete coordinates, similar to Kosmos~\cite{Kosmos}. These bounding boxes are predicted in a canonical reading order, which is described further in the supplementary material. The following regular expression shows the output format for each box in the maximal-information setting:

\begin{small}
\begin{center}
\sred{<x\_(\string\d+)>}%
\sred{<y\_(\string\d+)>}%
\sgreen{(.*?)}%
\sblue{<x\_(\string\d+)>}%
\sblue{<y\_(\string\d+)>}\newline
\spurple{<class\_([\string^>]+)>}
\end{center}
\end{small}

\noindent where the the \sred{first group} denotes the coordinates of the top-left corner, the \sgreen{second group} denotes the text contained within the bounding box, the \sblue{third group} denotes the coordinates of the bottom-right corner, and the \spurple{final group} represents the semantic class.

Note that each of these groups is optional and their presence in the model's output for a given image would depend on the prompt combination specified for that sample.

We adopt the tokenizer used by Taylor et al.~\cite{taylor2022galacticalargelanguagemodel}, as their model is also specialized for the scientific text domain. The coordinates of the bounding boxes, the semantic classes and the seven prompt components are all added as dedicated special tokens. This adds $H + W + C + 7$ tokens in total to the tokenizer vocabulary, where $C$ is the number of semantic classes.

\subsection{Datasets}

Compared to existing methods, such as Kosmos-2.5~\cite{Kosmos} and Nougat~\cite{Nougat}, \eclair is trained on a relatively smaller dataset as summarized in Table~\ref{tab:training-datasets}.

The arXiv-5M dataset makes up a large portion of our training data and it supports the maximum-information prompt (MIP) described in Section~\ref{sec:prompt}. The generation pipeline used to create this dataset is discussed further in Section~\ref{sec:latex_to_pdf}. We pre-train \eclair on this dataset.

We find that recent models such as Nougat~\cite{Nougat}, that are only trained on academic documents, do not handle visually-diverse documents very well, often either degenerating into hallucinations or repetition loops or simply terminating early by predicting the end-of-sequence token. We hypothesize that this is because the training data lacks the heterogeneity needed to handle more complex layouts such as magazines, leaflets, and picture-books. To address this, we fine-tune \eclair further on the arXiv-5M along with several publicly available datasets with diverse layouts and domains, such as DocLayNet~\cite{Doclaynet}, SynthTabNet~\cite{nassar2022tableformer} and G1000~\cite{g1000}. We also create a high-quality human-annotated dataset consisting of documents sampled from the Common Crawl corpus~\cite{commoncrawl}. Additionally, we create a README dataset by sampling README documents from the Stack~\cite{Kocetkov2022TheStack} and rendering them using Pandoc~\cite{pandoc}. Most of these datasets contain only partial annotations and the maximum information available in each is summarized in Table~\ref{tab:training-datasets}. The pre-processing steps for these datasets are described in more detail in the supplementary material.
\begin{table}[t!]
    \centering
    \footnotesize
    \begin{tabular}{lcc}
        \toprule
        \textbf{Dataset}         & \textbf{Size}   & \textbf{Modality} \\
        \midrule
        \textbf{arXiv-5M}        & 5M              & {Structured, Boxes, Classes}\\\hline
        \textbf{SynthTabNet~\cite{nassar2022tableformer}}     & 480K            & {Structured, Boxes, Classes}\\\hline
        \textbf{README}   & 302K            & {Structured} \\\hline
        \textbf{DocLayNet~\cite{Doclaynet}}       & 56K             & {Plain, Boxes, Classes}\\\hline
        \textbf{G1000~\cite{g1000}}      & 324K            & {Plain}\\\hline
        \makecell[l]{\textbf{Human-labeled} \\
        \textbf{Common Crawl} \\ 
        \textbf{samples}} & 14K             & {Plain, Boxes, Classes}\\
        \bottomrule
        \textbf{Total} & 6.176M & \\

    \end{tabular}
    \vspace{-3mm}
    \caption{Summary of the datasets used to train \eclair, including a description of the maximum information available in the annotations of each dataset. \vspace{-12pt}}
    \label{tab:training-datasets}
\end{table}

\subsection{The arXiv-5M Dataset}~\label{sec:latex_to_pdf}

In the introduction, we briefly discussed the need for a dataset that provides labels for our maximum information setting, i.e. bounding boxes, semantic classes and formatted text with formulas and tables, all in reading order. Since no such dataset exists, we created a new one.

Our approach is inspired by Nougat~\cite{Nougat}, where the authors create ground truth image/markdown pairs from arXiv papers. Their pipeline relies on LatexML, a tool to convert {\LaTeX} source code to HTML, which they convert to markdown subsequently. We follow a different approach, which handles both the {\LaTeX} compilation and the conversion to structured output at the same time (instead of using separate processing pipelines for each) and hence retains the relationship between text and image down to character-level bounding boxes and allows us to extract semantic classes for each box.
Our representation for the structured text output
\begin{itemize}
    \item consists of rectangular boxes
    \item has a semantic class assigned to each box
    \item represents normal text and formatted text as markdown
    \item represents tables and formulas as {\LaTeX}
\end{itemize}
The box and class information can be used to re-arrange the order of content (e.g. footnotes at the end) and to filter unwanted content, for example page headers and footers.

We modify the open-source {\TeX} Live distribution by adding hooks inside the {\TeX} compiler itself and embedding a Python interpreter for further processing on-the-fly. We hook the internal {\TeX} methods for node, character and hbox/vbox allocations, token reading and output generation and forward these to a custom Python class that keeps track of the elements from allocation to output on the PDF page. Multiple stacks are used to keep track of how the elements are nested in the input and output, and a rule-based system generates a nested hierarchy with the elements of interest.

With this method we generated a high-quality ground-truth dataset consisting of around roughly 5 million pages which we call arXiv-5M.

\section{Results}
The details about our experimental setup and training strategy can be found in the supplementary material. 

\vspace*{-\topskip}:
\begin{table*}[t]
    \centering
  \begin{tabular}{l|c|cccccccc}
\toprule
Method & \shortstack{Mask\\out} &\shortstack{Counting\\F1**}↑ & WER ↓ & \shortstack{Edit\\distance} ↓ & F1 ↑ & Precision ↑ & Recall ↑ & BLEU ↑ & METEOR ↑ \\
\midrule
{\shortstack{ÉCLAIR-MIP}}& \ballotx & 0.934 & \textbf{0.142} & 0.109 & \textbf{0.942} & 0.960 & 0.942 & \textbf{0.886} & \textbf{0.930} \\ \hline
{\shortstack{\vspace{-2pt}\\ÉCLAIR-MIP}}& \ballotcheck & \textbf{0.937} & 0.146 & \textbf{0.108} & 0.941 & \textbf{0.966} & 0.936 & 0.885 & 0.927 \\ \hline
{\shortstack{\vspace{-2pt}\\Kosmos-2.5\\(ocr-mode)}}& \ballotcheck  & 0.919 & 0.195 & 0.114 & 0.937 & 0.932 & \textbf{0.950} & 0.862 & 0.927 \\ \hline
{\shortstack{\vspace{-2pt}\\Kosmos-2.5\\(md-mode)}}& \ballotcheck  & 0.843 & 0.249 & 0.184 & 0.890 & 0.941 & 0.876 & 0.805 & 0.851 \\ \hline
{\shortstack{\vspace{-2pt}\\GOT\\(ocr-mode)}}& \ballotcheck  & 0.776 & 0.302 & 0.216 & 0.818 & 0.863 & 0.825 & 0.713 & 0.795 \\ \hline
{\shortstack{\vspace{-2pt}\\GOT\\(md-mode)}} & \ballotcheck  & 0.825 & 0.259 & 0.157 & 0.879 & 0.908 & 0.875 & 0.760 & 0.852 \\
\bottomrule
\end{tabular}
\vspace{-2mm}
    \caption{Evaluation results on \benchds. Reported standard NLTK metrics \cite{nltk} are character level (Edit-distance) or word level (F1, Precision, Recall, BLEU, METEOR) metrics typically used by the OCR and natural language processing (NLP) communities. We also report Counting F1 and word error rate/word edit distance metrics. \\
    *MIP-maximal-information prompt\\
    **Counting F1 score is computed over the set $\{$ $\texttt{he}_1$, $\texttt{said}_1$, $\texttt{that}_1$, $\texttt{she}_1$, $\texttt{said}_2$, $\texttt{that}_2$, $\texttt{they}_1$, $\texttt{said}_3$, $\texttt{that}_3$, $\texttt{he}_2$, $\texttt{said}_4$, $\texttt{something}_1$ $\}$. This allows to track and penalize words that missed but has more than one occurrence in the document. \vspace{-10pt}}
    \label{tab:eclair-results}
\end{table*}

\subsection{Reading Order benchmark}
\label{ro_bench}

\begin{figure}[h]
\centering
\includegraphics[width=\linewidth]{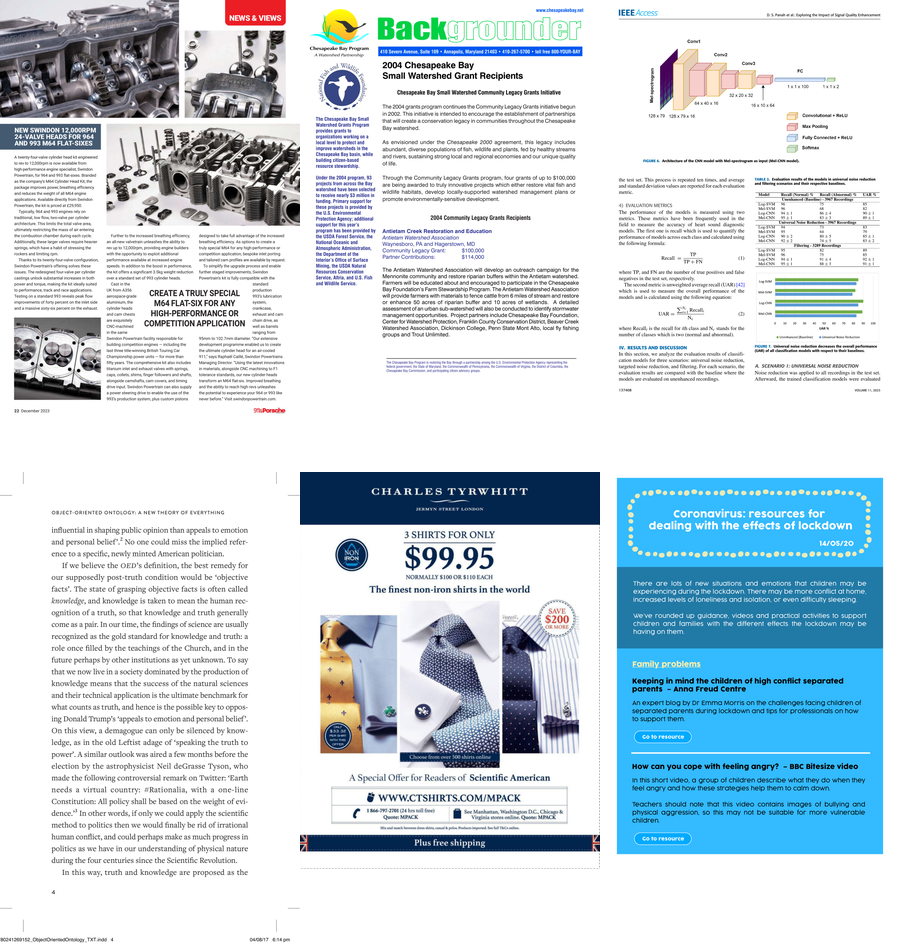}
\vspace{-3mm}
\caption{Example pages from \benchds, our visually diverse document benchmark.}
\label{fig:test-set-examples}
\vspace{-4mm}
\end{figure}

\textbf{{\benchds} Evaluation.} We evaluate the reading order accuracy of \eclair against known SOTA methods like Kosmos-2.5 \cite{Kosmos} and GOT \cite{GOT}. Both of these methods have two output modalities - a plain OCR mode and a markdown mode, and we compare \eclair with both modes. For this evaluation, we utilize an internally curated and human-labeled diverse set of PDFs, comprising a total of 789 pages sampled from various sources such as magazines, books, and the Common Crawl corpus \cite{sebastian_spiegler_statistics_2013} which we call \benchds (Document Reading Order, Bounding boxes, and Semantic classes), see Figure~\ref{fig:test-set-examples}. This approach aims to cover a diversity of layouts similar to those found in DocLayNet \cite{Doclaynet}. We instructed human annotators to provide annotations on this dataset following the same labeling system as DocLayNet due to its comprehensive human annotation guidelines. However, we added additional requirements, the most significant being the inclusion of reading order. We will make the selected pages and associated annotations available to the research community to serve as an additional and complementary benchmark for document understanding and OCR. 

Prior to evaluation, we perform three preprocessing steps on the predictions and corresponding ground truth labels. First, we apply string normalization \cite{string-normalize} to remove all non-alphanumeric characters, convert sequences of whitespaces to a single space, and strip any leading or trailing whitespaces to ensure fair comparison. Second, to address variations in the output formats for tables and equations among different methods, and given that our current evaluation benchmark does not include labels for equations and tables, we mask out these elements in the images used for inference across all considered methods. Additionally, for GOT (md), we also mask-out headers and footers from the images, as the model seems to ignore these elements. 
Lastly, we also filter out {\TeX} commands present in GOT prediction e.g. for title, section and sub-section headers, since those would otherwise penalize the model in text-only metrics.

The results in Table~\ref{tab:eclair-results} show how \eclair outperforms both Kosmos and GOT on most metrics. Since these models are trained with multiple prompt modes similar to \eclair, we compare the output of \eclair in MIP mode against Kosmos-2.5 and GOT in both OCR and MD modes. We observe that Kosmos-2.5 performs better in OCR mode, where it produces bounding boxes; however, GOT exhibits the opposite behavior, performing better in MD mode as opposed to OCR mode. We believe this discrepancy is due to differences in data blending during training. \eclair in MIP mode produces both MD for text inside the bounding boxes and shows superior accuracy compared to both other methods.

Since there is no common validation set available for comparing equation and table extraction across methods, we evaluate our formula and table extraction accuracy on a validation set derived from arXiv (See Section~\ref{sec:arxiv}).

\begin{table*}[t]
\scriptsize
\centering
\setlength{\tabcolsep}{3pt}
\renewcommand{\arraystretch}{1.1}
\resizebox{\textwidth}{!}{
\begin{tabular}{l|c|cccccccccccc}
\toprule[.9pt]
\multirow{2}{*}{Method} & \multirow{2}{*}{Size} & \multicolumn{2}{c}{Edit Distance$\downarrow$} & \multicolumn{2}{c}{F1-score$\uparrow$}  &  \multicolumn{2}{c}{Precision$\uparrow$} & \multicolumn{2}{c}{Recall$\uparrow$} & \multicolumn{2}{c}{BLEU$\uparrow$} & \multicolumn{2}{c}{METEOR$\uparrow$}  \\  
\cmidrule(rl){3-4} \cmidrule(rl){5-6}  \cmidrule(rl){7-8}  \cmidrule(rl){9-10}  \cmidrule(rl){11-12} \cmidrule(rl){13-14} 
& & en & zh & en & zh & en & zh & en & zh & en & zh & en & zh \\ 
\midrule
Nougat~\cite{Nougat} & 250M & 0.255 & - & 0.745 & - & 0.720 & - & 0.809 & - & 0.665 & - & 0.761 & -\\ 
TextMonkey~\cite{liu2024textmonkey} & 7B & 0.265 & - & 0.821 & - & 0.778 & - & 0.906 & - & 0.671 & - & 0.762 & -\\
DocOwl1.5~\cite{hu2024mplugdocowl1.5} & 7B & 0.258 & - & 0.862 & - & 0.835 & - & 0.962 & - & 0.788 & - & 0.858 & - \\
Vary~\cite{wei2023varyscalingvisionvocabulary} & 7B & 0.092 & 0.113 & 0.918 & 0.952 & 0.906 & 0.961 & 0.956 & 0.944 & 0.885 & 0.754 & 0.926 & 0.873 \\
Vary-toy~\cite{wei2024smalllanguagemodelmeets} & 1.8B & 0.082 & 0.142 & 0.924 & 0.914 & 0.919 & 0.928 & 0.938 & 0.907 & 0.889 & 0.718 & 0.929 & 0.832 \\  
Qwen-VL-Plus~\cite{Qwen-VL} & - & 0.096 & 0.121 & 0.931 & 0.895 & 0.921 & 0.903 & 0.950 & 0.890 & 0.893 & 0.684 & 0.936 & 0.828 \\
Qwen-VL-Max~\cite{Qwen-VL} & 72B+ & 0.057 & 0.091 & 0.964 & 0.931 & 0.955 & 0.917 & 0.977 & 0.946 & 0.942 & 0.756 & 0.971 & 0.885 \\ 
Fox~\cite{liu2024focus_fox} & 1.8B & 0.046 & 0.061 & 0.952  & 0.954 & 0.957 & 0.964 & 0.948 & 0.946 & 0.930 & 0.842 & 0.954 & 0.908 \\ 
GOT~\cite{GOT} & 580M & 0.035 & 0.038 & \textbf{0.972} & 0.980 & \textbf{0.971} & 0.982 & 0.973 & 0.978 & 0.947 & 0.878 & 0.958 & 0.939 \\
\textbf{\eclair} & 936M & \textbf{0.032} & - & 0.968 & - & 0.962 & - & \textbf{0.974} & - & \textbf{0.950} & - & \textbf{0.980} & - \\
\bottomrule[.9pt]
\end{tabular}
}
\vspace{-2mm}
\caption{Accuracy comparison of various methods across different metrics in both English and Chinese (zh). Currently \eclair doesn’t train with additional chinese data or other form of multi-lingual data. The numbers in top row are obtained from GOT~\cite{GOT}.}
\label{ref-table-got}
\end{table*}

\begin{table*}[h]
\centering
    \begin{tabular}{l|c|c|cccccc}
    \toprule[.9pt]
    {Method} &{Size}  &{Modality} & {Edit Distance$\downarrow$} & {BLEU$\uparrow$} & {METEOR$\uparrow$} & {Precision$\uparrow$} & {Recall$\uparrow$} & {F1-score$\uparrow$} \\
    \midrule
    \multirow{4}{*}{{Nougat-Base}~\cite{Nougat}} & \multirow{4}{*}{350M} & All & 0.071 & 0.891 & 0.930 & 0.935 & 0.928 & 0.931 \\
     & & Text   & 0.058   & 0.912 & 0.946 & 0.962 & 0.953 & 0.957 \\
    
    & & Math  & 0.128   & 0.569 & 0.754 & 0.765 & 0.766 & 0.765 \\ 
    
     &  &Tables & 0.211   & 0.697 & 0.791 & 0.754 & 0.807 & 0.780 \\
    \midrule
    \multirow{4}{*}{{\eclair}} & \multirow{4}{*}{963M} & All & {0.026 }& {0.952} & {0.998} & {0.970} & { 0.970} & {0.970}     \\
     & &  Text       & {0.015}      & {0.979}     & {0.996}     & { 0.992}     & {0.990 }    & 0.990     \\
     &  & Math  & {0.123}    & {0.679 }& 0.934    & 0.858 & {0.860 } & 0.853    \\
     &  & Tables     & {0.064}      & {0.871 }    & {0.992}     & { 0.918   }  & {0.916}     & {0.916} \\
    \bottomrule[.9pt]
    \end{tabular}
    \vspace{-2mm}
    \caption{Evaluation of the Nougat-Base model on the Nougat validation set (as reported in ~\cite{Nougat}), and of \eclair pre-trained on the arXiv-5M dataset and validated on the corresponding validation set. \textbf{Note}: We do not aim to provide a direct comparison between Nougat and \eclair here due to the concerns discussed in Section~\ref{sec:arxiv}. \vspace{-10pt}}
    \label{tab:eclair_maths}
\end{table*}

\textbf{Training \& Inference Ablation.} All the results for \eclair presented in Tables~\ref{tab:eclair-results} and~\ref{ref-table-got} were obtained with a repetition penalty~\cite{keskar2019ctrl} of $1.1$ applied during inference. Table~\ref{tab:ds_inf_ablation} demonstrates the value of this inference-time hyperparameter and of the additional datasets added in the fine-tuning stage.

\textbf{GOT Benchmark.} Along with \benchds, we evaluate \eclair on the GOT benchmark proposed in Fox~\cite{liu2024focusfinegrainedmultipagedocument}, with the results shown in Table~\ref{ref-table-got}. \eclair with a size of 936M parameters and in MIP mode demonstrates competitive or superior accuracy across most metrics, particularly excelling in Edit Distance and Recall, where it achieves the best scores among the models compared. Notably, \eclair outperforms several larger models, such as Qwen-VL-Max (72B)~\cite{Qwen-VL}, Fox (1.8B)~\cite{liu2024focusfinegrainedmultipagedocument}, despite having a significantly smaller parameter size. While GOT (580M)~\cite{GOT} and \eclair exhibit similar accuracy, \eclair employs a decoder that is approximately half the size of GOT's decoder.

\begin{table}[]
   \centering
   \begin{tabular}{c|c|c|cc}
       \toprule 
        \shortstack{Pre\\Training} &  \shortstack{Fine\\Tuning} & \shortstack{Repetition\\Penalty$=1.1$} &\shortstack{Counting\\F1**}↑ & WER ↓ \\ \midrule
        \ballotcheck & \ballotx & \ballotx & 0.663 & 0.264 \\ 
        \ballotcheck & \ballotcheck & \ballotx & 0.925 & 0.151 \\ 
        \ballotcheck & \ballotcheck & \ballotcheck & 0.934 & 0.142 \\ \bottomrule
   \end{tabular}
   \vspace{-2mm}
   \caption{Comparison of \eclair on \benchds before and after the fine-tuning stage, and also with and without a repetition-penalty (after the fine-tuning stage). \vspace{-10pt}}   
   \label{tab:ds_inf_ablation}
\end{table}
     
\subsection{Extraction of Formulas and Tables}~\label{sec:arxiv}

In this section, we evaluate the extraction quality of \eclair on some important semantic-classes: formulae, tables and text. The latter consists of all semantic classes excluding formulae and tables. We report our findings on the validation set (10,000 samples) associated with the arXiv-5M dataset in Table~\ref{tab:eclair_maths}. Our results demonstrate good extraction quality overall, with table and math elements being harder for \eclair to transcribe compared to text elements. Note that these metrics are reported for \eclair pre-trained on arXiv-5M (i.e., prior to fine-tuning).

Since other methods such as Nougat~\cite{Nougat} cannot be directly compared to \eclair on our validation set owing to non-trivial differences in their output formatting styles, we cannot provide a direct comparison here. However, since Nougat and the pre-trained \eclair model are both trained on academic documents and evaluated on data sampled from arXiv, we find it useful to present the extraction quality of Nougat on these categories on their own validation set as a point of reference. These results are also summarized in Table~\ref{tab:eclair_maths}. We observe similar trends in Nougat's extraction quality for math and table elements, as discussed above.

\subsection{Document Object Detection}

 We evaluate the accuracy of detection of semantic text blocks of \eclair on the DocLayNet benchmark. Following~\cite{swindocsegmenter}, we fine-tune \eclair solely on DocLayNet for 50k steps to ensure that the bounding box class labels are not biased by labeling styles of other datasets (such as merging of several header and footer boxes). In order to compare to SOTA methods that report coco-mAP, we report the same metric using class token logits for ranking the predicted bounding boxes. We note, however, that being an autoregressive generator, \eclair remains in inherent disadvantage on coco-mAP metric compared to standard detectors due to it predicting bounding boxes and classes inline with the text in reading order, leading to inability of overprediction to 100 bounding boxes assumed by coco-mAP. On the other hand, this results in \eclair not requiring non-maximum-suppression or threshold selection. For this reason, previous autoregressive object detectors adopt various tricks to improve the recall at low precision~\cite{pix2seq}. We also follow this approach and adopt sequence augmentation \cite{pix2seq} with noisy and duplicate bounding boxes as well as sampling of top-k class labels from each predicted bounding box during inference for reporting coco-mAP. The comparison with SOTA methods is presented in Table \ref{tab:eclair_doclay}, where we compare to reported Mask R-CNN~\cite{he2017mask} metrics and reproduced SwinDocSegmenter~\cite{swindocsegmenter}. As can be seen, \eclair is competitive even compared to specialized object detectors. 
 
 Nevertheless, in agreement with previous works on autoregressive detection \cite{avetisyan2024scenescriptreconstructingscenesautoregressive}, we find mAP to be a suboptimal metric for such scenario. We provide further discussion on the evaluation metrics in the supplementary material, with more detailed evaluation of \eclair and competing methods. 
 \begin{table}[h]
    \centering
    \centering
\setlength{\tabcolsep}{5pt}
    \begin{tabular}{l|ccc}
        \toprule
        Classes        & \shortstack{Mask-RCNN\\\cite{he2017mask}} & \shortstack{SwinDoc\\Segmenter\cite{swindocsegmenter}}  & \eclair  \\ \midrule
        Caption        & 71.5 & 83.5 & \textbf{83.5} \\ 
        Footnote       & \textbf{71.8} & 67.8 & 66.9 \\ 
        Formula        & 63.4 & 64.2 & \textbf{65.7} \\
        List-item      & 80.8 & \textbf{84.1} & 79.0 \\ 
        Page-footer    & 59.3 & \textbf{65.1} & 62.0 \\ 
        Page-header    & 70.0 & \textbf{71.3} & 70.7 \\ 
        Picture        & 72.7 & \textbf{85.6} & 76.9 \\ 
        Sec-header     & \textbf{69.3} & 68.0 & 67.0 \\ 
        Table          & 82.9 & \textbf{86.0} & 77.6 \\ 
        Text           & \textbf{85.8} & 84.5 & 82.0 \\
        Title          & 80.4 & 66.8 & \textbf{82.0} \\ \midrule
        All            & 73.5 & \textbf{75.2} & 73.9 \\ \bottomrule
    \end{tabular}
    \vspace{-2mm}
    \caption{COCO-mAP (with defaults IoU=0.5:0.95, area=all, maxDets=100) on DocLayNet Benchmark. \vspace{-15pt}}
    \label{tab:eclair_doclay}
\end{table}

\subsection{LLM Benchmark}
\eclair enables content extraction from PDFs, PPTs, and other scanned documents to meet the growing demands for high-quality data to train large language models (LLMs) \cite{parmar2024nemotron415btechnicalreport, nvidia2024nemotron4340btechnicalreport, dubey2024llama3herdmodels, gemmateam2024gemma2improvingopen}. Unlike conventional extraction tools, e.g., PyMuPDF4LLM~\cite{pymupdf4llm}, \eclair is engineered to preserve semantic integrity and textual coherence. In this section, we compare the effectiveness of \eclair and PyMuPDF4LLM~\cite{pymupdf4llm} for this task. We do this by training the Nemotron-8B LLM model from scratch on the text extracted by both of these methods from a common set of PDF documents, and compare the trained models on the Massive Multitask Language Understanding (MMLU)~\cite{hendrycks2020measuring} benchmark, an average of multiple other benchmark scores including: ARC-Easy and ARC-Challenge~\cite{clark2018think}, HellaSwag~\cite{zellers2019hellaswag}, OpenBooxQA~\cite{mihaylov2018can}, PIQA~\cite{bisk2020piqa}, RACE~\cite{lai2017race}, WinoGrande~\cite{sakaguchi2021winogrande}, TriviaQA~\cite{Joshi2017TriviaQAAL}. The results of this experiment, summarized in Table~\ref{tab:eclair_llm}, highlight \eclair’s effectiveness in extracting high quality training data for improved LLM accuracy. Details about the training setup and post-processing steps for \eclair can be found in the supplementary material. 
\begin{table}[h]
    \centering
\setlength{\tabcolsep}{4pt}
    \begin{tabular}{l|c|cc}
        \toprule
        Method  & \shortstack{Tokens \\Extracted \\(B)} & MMLU  $\uparrow$ & \shortstack{Other\\ Bench \\Avg} $\uparrow$ \\
        \hline
        PyMuPDF4LLM~\cite{pymupdf4llm} & 43.6 & 37.2 & 55.72 \\
        \eclair & 55.1            & \textbf{39.1} & \textbf{56.7} \\
        
        \bottomrule
    \end{tabular}
    \vspace{-2mm}
    \caption{Comparison of the Nemotron-8B accuracy when trained on data extracted with \eclair or PyMuPDF4LLM~\cite{pymupdf4llm}. \vspace{-10pt}}
    \label{tab:eclair_llm}
\end{table}

\subsection{Multi-token Inference}
\label{multi_token_inference}
An important shortcoming of autoregressive models, including those targeted at OCR applications, is the large number of decoding steps necessary for text extraction, resulting in slow inference speed. In a standard autoregressive decoding formulation, each subsequent $l^{th}$ token in the sequence is decoded incrementally, based on the context of $t_0:t_{l-1}$ tokens. For text-dense images, such as documents, this results in a large amount of decoding steps, at least equal to the number of tokens in the sequence. 

To mitigate this, we investigate multi-token generation as an alternative inference method. Instead of next-token prediction, we train \eclair to predict $n$ subsequent tokens at a single step, and therefore reduce the number of necessary decoding steps by a factor of $n$. 

Specifically, for predicting $n$ tokens simultaneously, during training we introduce $n-1$ new linear layers on top of the final hidden state of the decoder. The output of each of these linear layers is subsequently input into the shared decoder head. During training, standard teacher forcing is applied, with next $n$ tokens representing the groundtruth for each corresponding context. During inference, we greedily decode the sequence $n$ tokens at a time. We do not perform token verification \cite{cai2024medusa} but rely on purely greedy decoding in the interest of maximal throughput at batch-inference.

Following prior work \cite{betterfaster}, we additionally perform experiments where we only consider the first predicted token during inference while the rest are discarded, while during training next-n tokens are predicted. In other words, we evaluate $n$-token trained models at next-token prediction. 

We experimented with 2-, 3-, or 4-token prediction and the results are reported in Table~\ref{tab:multi-token-results}, where $\frac{tkn}{step}$ corresponds to the number of tokens kept from multi-token prediction at a single decoding step during inference. As can be seen, multi-token \eclair is matching or outperforming the baseline at 2 or 3 tokens, which is equivalent to around 2x inference speed increase. At 4 tokens, the accuracy degrades. We nevertheless find that keeping only the first token during inference is a valid strategy for improving OCR metrics for any of the variants. We additionally report the inference speed of multitoken \eclair as well as competing methods, with average time per image on \benchds test set. As per-image speed of each method is related to its text extraction capabilities (due to the possibility of hallucination loops), we also report the average time per 100 tokens. As can be seen, multi-token approach allows \eclair to outperform competing methods speed-wise, despite it being a bigger model parameter-wise. The details on evaluation protocol can be found in the supplementary material.
\begin{table}[h]
\centering
\setlength{\tabcolsep}{5pt}
\begin{tabular}{l|c|cccc}
\toprule
                  {Method}                      &       {$\frac{tkn}{step}$}            & {WER} $\downarrow$  & {F1} $\uparrow$ & $\frac{sec}{img}$ $\downarrow$ & $\frac{sec}{100}$ $\downarrow$  \\ \midrule
                  Nougat \cite{Nougat} & 1 & - & - & 4.7 & 0.41 \\ 
                  GOT \cite{GOT} & 1 & 0.25 & 0.82 & 9.8 & 0.90 \\ 

{\eclair}                          &      1             & 0.14          & 0.93       & 3.8 & 0.42  \\ \midrule
\multirow{2}{*}{{\eclair}-2tkn} & 2 & 0.13          & 0.94  & 2.5 & 0.31 \\
                                             & 1 & \textbf{0.12} & \textbf{0.95} & 3.8 & 0.42\\
                                         \midrule
\multirow{2}{*}{{\eclair}-3tkn} & 3 & 0.15          & 0.92   & 1.77 & 0.23       \\
                                       & 1 &      0.13         &    0.94   & 3.8 & 0.42        \\
                                         \midrule
\multirow{2}{*}{{\eclair}-4tkn} & 4 & 0.17          & 0.90    & \textbf{1.32} & \textbf{0.20}   \\
                                         & 1 & 0.14          & 0.94   & 3.8 & 0.42 \\   
                                         \bottomrule
\end{tabular}
\vspace{-2mm}
\caption{Results and speed of multi-token models and competing methods. We report the average speed per image on \benchds test set ($\frac{sec}{img}$), and speed per 100 tokens ($\frac{sec}{100}$). These values are obtained from a PyTorch-based inference pipeline on an NVIDIA H-100 GPU. \vspace{-15pt}}\label{tab:multi-token-results}
\end{table}

\section{Related Work}
\textbf{Document Understanding Models} Models like LayoutLMv3 \cite{layoutlm3} excel in parsing complex documents for tasks such as layout analysis and visual question answering. However, they rely heavily on pre-extracted text, images, and bounding boxes, forming a brittle pipeline that can be error-prone due to its dependence on external systems. SwinDocSegmenter \cite{swindocsegmenter} and specialized variants of YOLO \cite{yolov5} have been trained for document-specific detection tasks without requiring additional inputs. While they effectively detect objects within documents, they generally do not output any text associated with these objects, lacking integrated OCR capabilities.

\textbf{Object Detection in Documents} is crucial for identifying and localizing elements within documents, aiding tasks like OCR and determining reading order. Traditional models such as Faster R-CNN \cite{ren2016faster} and Mask R-CNN \cite{he2017mask} have been adapted for document analysis, effectively detecting and segmenting components like text blocks, images, and tables. Despite their success, these models typically do not provide textual content alongside the detected objects, limiting their usefulness for comprehensive document understanding.

\textbf{End-to-End OCR-Free Models} that do not depend on external OCR systems have gained attention. Donut \cite{donut} introduced a transformer-based encoder-decoder architecture pre-trained on general text documents. Building on this, Nougat \cite{Nougat} extended training to scientific documents, outputting structured markdown with \LaTeX\ tables and equations. GOT \cite{GOT} focused on enhancing the recognition of specialized documents containing molecular formulas, sheet music, and charts. Kosmos-2.5 \cite{Kosmos} incorporated both markdown and plain text data with bounding boxes, introducing a prompt structure that allows users to choose between different output formats. However, these models may require compromises in prompt structures or may not handle a wide variety of document layouts effectively. Our proposed model, \eclair, is specifically trained to handle a greater variety of document layouts without requiring compromises in the prompt structure.

\textbf{Multimodal Large Language Models} like QwenVL~\cite{Qwen-VL}, GPT-4O~\cite{openai2024gpt4o} and Claude~\cite{anthropic2024claude} have demonstrated impressive OCR and document understanding capabilities, including the extraction of complex equations and tables in structured formats. While powerful, these models are large and computationally expensive, making them impractical for scaling to millions of pages. In contrast, \eclair\ is a sub-1B parameter model optimized for inference speed with multi-token decoding. 

\section{Conclusion}

In this work, we have presented \eclair, a general-purpose end-to-end text-extraction model. \eclair is the first model that extracts structured text, bounding boxes and semantic classes all at the same time. We are releasing a new benchmark dataset {\benchds} to capture the variety of layouts of various online documents and have shown that \eclair outperforms all current competitors on this benchmark. Additionally, we investigate and provide a technique to improve the inference time for \eclair. We hope that this will aid the OCR community in improving document-based text extraction, and benefit the LLM community by increasing the availability of previously unseen text data for training.

\section{Acknowledgments}
We would like to thank Shrimai Prabhumoye, John Kamalu, Brandon Norick, Mostofa Patwary, Thomas Breuel, Osvald Nitski, Ushnish De, Mehrzad Samadi, Guilin Liu, Zhiding Yu, Mike Ranzinger, Teo Ene, Mohammad Shoeybi and Bryan Catanzaro, for their feedback and valuable discussions.

{\small
\bibliographystyle{ieee_fullname}
\bibliography{egbib}
}

\clearpage 
\setcounter{section}{0} 
\setcounter{figure}{0}  
\setcounter{table}{0}   

\renewcommand{\thesection}{S\arabic{section}} 
\renewcommand{\thefigure}{S\arabic{figure}}   
\renewcommand{\thetable}{S\arabic{table}}     

\section*{Supplementary Material} 
\addcontentsline{toc}{section}{Supplementary Material} 

\section{Architecture Details}

The entire architecture has a total of 937M parameters. 

\textbf{Vision Encoder}  Our vision encoder is initialized from the RADIO~\cite{radio} ViT-H /16 model (657M parameters). The inputs to the encoder are images of resolution $1280\times 1024$. We resize and pad the original image to this size, while maintaining the aspect ratio. This results in $80\times 64$ patches. 

\textbf{Vision Neck} Following the observation in \cite{hu2024mplugdocowl15unifiedstructure} that text is more correlated within a line rather than across lines, we compress the sequence with a horizontal convolutional kernel of size $1\times 4$ and stride $1\times 4$. Given an input image of size $1280\times 1024$ which produces $5120$ patches, this reduces the sequence length to $1280$. 

\textbf{Decoder} We use an MBart~\cite{mbart} decoder with 10 layers (279M parameters) trained from scratch. The maximum sequence length for the decoder is set to 3584. 

\section{Training \& Inference}
We follow a two-step training strategy, similar to Nougat~\cite{Nougat}. We pre-train \eclair on the arXiv-5M dataset and fine-tune it further on all the training datasets. In both stages, we use the AdamW optimizer~\cite{loshchilov2017decoupled} and train for 130000 iterations with an effective batch size of 128. During pre-training, we use a constant learning rate of $2\times 10^{-5}$ with 5000 linear warmup steps. For fine-tuning, we employ a constant learning rate of $8\times 10^{-6}$ with 500 linear warmup steps. During inference, we use a greedy decoding strategy with a repetition penalty~\cite{keskar2019ctrl} of $1.1$. 

\textbf{Inference Speed Comparison} To measure the speed of competing methods compared to \eclair and its multi-token variants, we utilize their publicly shared huggingface pipelines and pre-trained models. We report the speed of exclusively the model excluding any pre-processing, i.e., \textit{model.generate()} step. All models are evaluated with the provided pre-processing pipelines and the recommended parameters are used. We perform evaluation in bf16 for fair comparison. Additionally, since the speed of a model is affected by potential hallucinations that can lead to infinite loops bound by maximal sequence length, we also report the speed per 100 tokens. We can see that in this case speed of \eclair is almost identical to that of Nougat, as their decoder architecture is the same. At the same time, while \eclair has the most parameters out of competing methods, being front-heavy, it enjoys faster inference speed which is primarily dictated by the size of the decoder.
For multi-token inference, we adopt a simple greedy decoding strategy (as opposed to more complex speculative strategies) for ensuring maximal throughput in batch mode during inference. Each linear layer has 1024 nodes.

\section{Datasets}
\subsection{Reading Order}

\begin{figure*}[h]
\centering
\includegraphics[width=\textwidth]{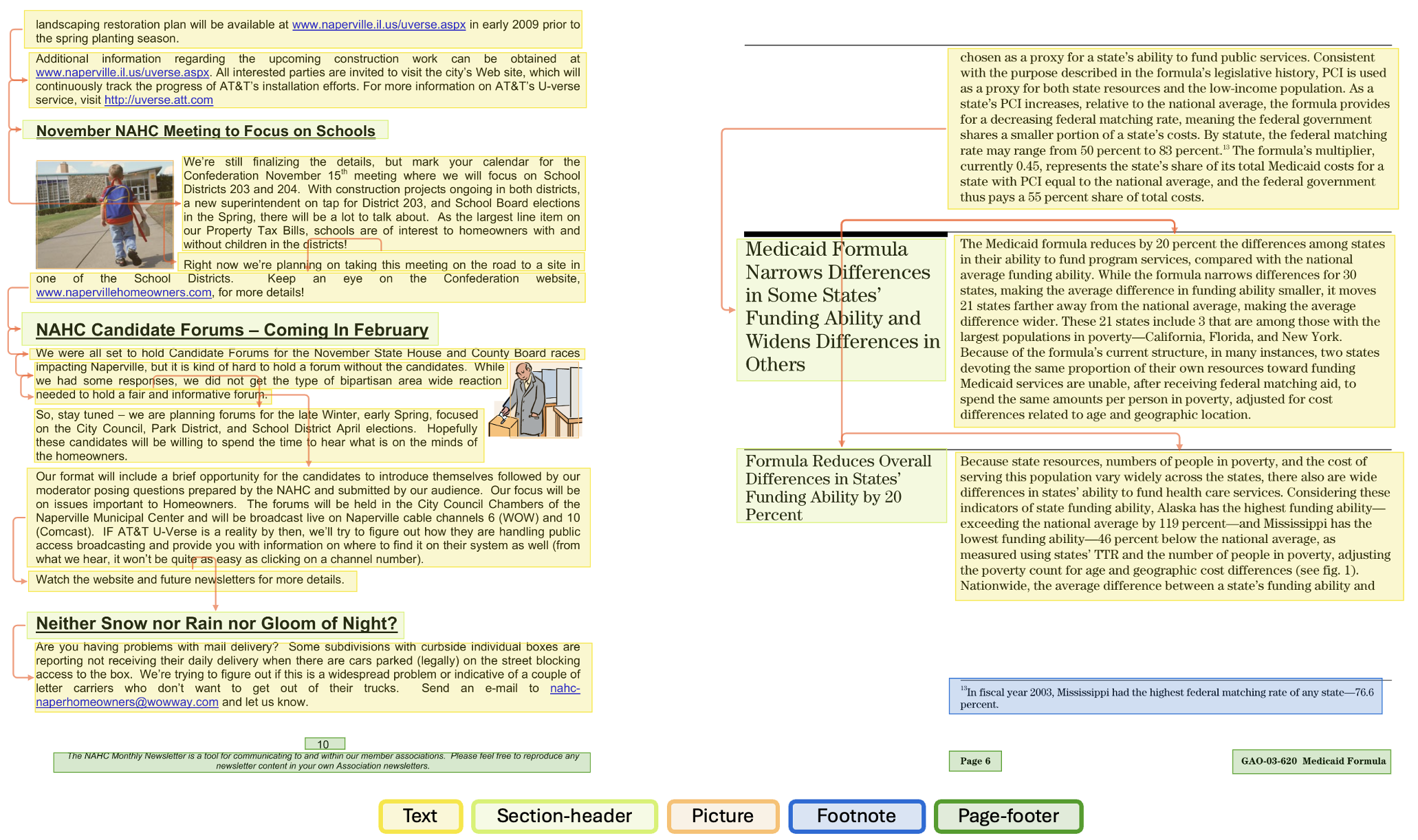}
\vspace{-3mm}
\caption{Illustrations of reading order over relevant text-like elements, i.e. Text, Section-header, List-item, Title and Formula. Other semantic classes (such as Picture, Footnote and Page-footer in the examples here) are not included in the reading order of the main body. (Note: We are not showing all the classes)}
\label{fig:reading-order}
\vspace{-4mm}
\end{figure*}

The reading order for the bounding boxes in a page is the order in which the contents of these boxes would be visited if a person was reading the page out aloud. We only include relevant text-like semantic classes in this order, i.e., Text, Section-Header, List-Item, Title and Formula. See Figure~\ref{fig:reading-order}  for an illustration. Additionally, we reorder all Page-Header elements to be at the start of the page, and all Footnote, Page-Footer, Picture, Table and Caption elements to be at the end.

\subsection{Pre-processing steps for the training datasets }

\textbf{SynthTabNet} The SynthTabNet dataset~\cite{nassar2022tableformer} consists of images of tabular data along with HTML annotations. We convert these HTML annotations to \LaTeX.

\textbf{G1000} We obtain the annotations for G1000 using Tesseract~\cite{smith2007overview}.

\textbf{README} We create a dataset of project documents by sampling README files from the Stack~\cite{Kocetkov2022TheStack}. To normalize the markdown source code, we convert it to HTML using Pandoc~\footnote{https://pandoc.org/} and render the converted HTML as PDF using wkhtmltopdf~\footnote{https://wkhtmltopdf.org/}. To obtain the ground truth for each page within the rendered PDFs, we first convert the HTML source back into our markup format and adopt Nougat's~\cite{Nougat} data processing pipeline\footnote{https://github.com/facebookresearch/nougat} to split and align the markdown with each page.

\textbf{Table Auto-Labeling} The DocLayNet dataset and the human-annotated CommonCrawl samples do not contain {\LaTeX} annotations for tables and equation blocks. When we mask these elements out (both in the input image and the targets) in the fine-tuning stage owing to lack of ground truth, we observe that the model tends to mis-classify tables as pictures during inference. We hypothesize that this is caused by the lack of tables in the visually diverse subset of the training data, rendering such samples out-of-distribution (OOD) during inference. We address this by fine-tuning \eclair on table crops from SynthTabNet and arXiv-5m and using it to auto-label table crops from DocLayNet and the human-annotated samples. 

\section{Post-processing: Hallucinations and Bad-box detection.}

Similar to Nougat~\cite{Nougat}, we observe that \eclair sometimes degenerates into repetition loops wherein it repeats the same phrase, sentence or paragraph over and over again at the end of the prediction. Nougat \cite{Nougat} detect hallucinations by tracking a moving average of logits and flagging the outputs when a certain threshold is reached.

\begin{figure*}[htbp]
    \centering
    \subfloat[]{\includegraphics[width=0.9\linewidth]{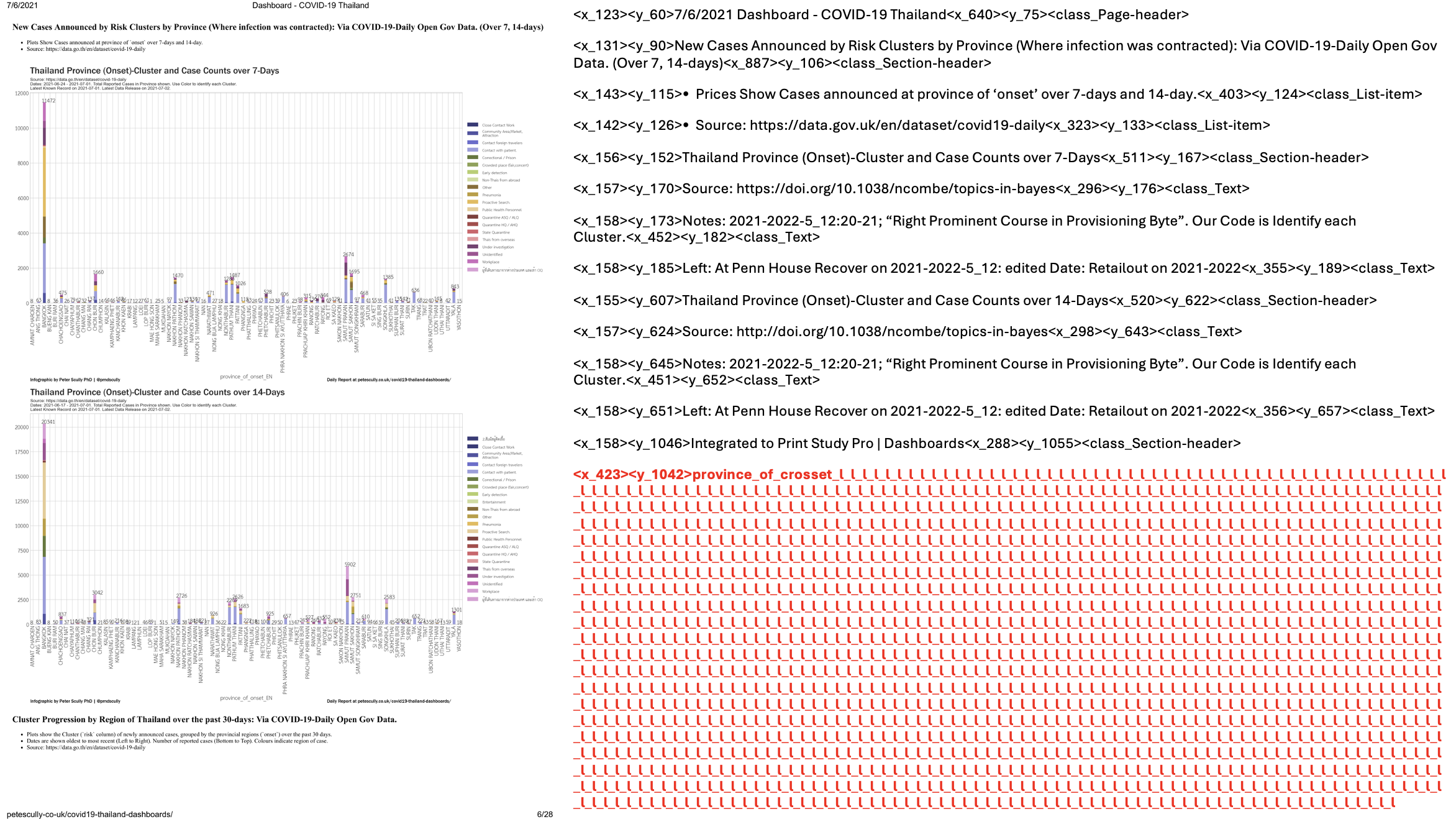}} \\
    \subfloat[]{\includegraphics[width=0.9\linewidth]{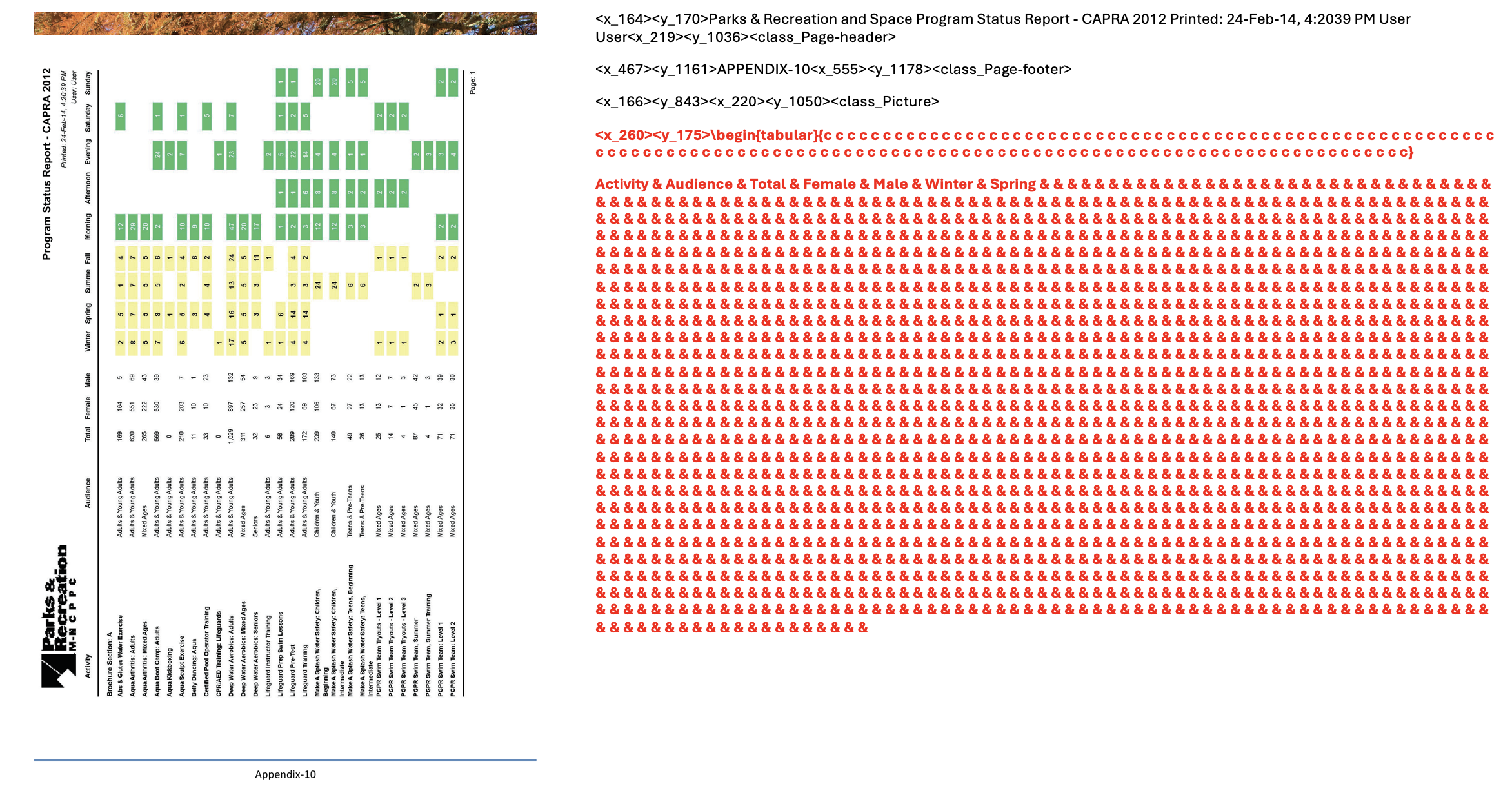}}
    \vspace{-3mm}
    \caption{Examples of hallucinations in the \eclair predictions. The hallucinations (in this case, repetition loops), marked in red, are detected and filtered out by our hallucination-mitigation strategy.}
    \vspace{-3mm}
    \label{fig:hallucinations}
\end{figure*}

For \eclair, we adopt a simple hallucination mitigation strategy to filter out such occurrences: the inference-time prompt is always set to the Maximal Informative Prompt (MIP) and we do a strict syntax check on the resulting predictions to reject non-compliant boxes. Some examples of hallucinations detected and filtered out using this strategy are shown in Figure~\ref{fig:hallucinations}. We also enforce the spatial and categorical validity of the remaining boxes by verifying that the bottom-right corner of each bounding box exceeds the top-left corner and that classes conform to a validated schema. By implementing this layered filtering strategy, we observe a substantial reduction in model hallucinations.

\section{Object detection}

\subsection{mAP - Back to Simple Metrics}
\label{sec:map}
In line with previous works on autoregressive detection \cite{avetisyan2024scenescriptreconstructingscenesautoregressive}, we find mAP to be inherently unfavorable to end-to-end models like \eclair. Dedicated object detectors generally predict a set of bounding boxes of a fixed size as a raw output, where each bounding box is associated with a confidence score. Consequently, it is possible to control the recall-precision trade-off of the model by adjusting the confidence threshold by which the raw predictions are filtered. Naturally, this results in a possibility of achieving a high recall for the model, albeit at low precision. Instead, \eclair predicts a set of bounding boxes in line in the output stream of tokens, requiring no filtering or thresholding.

An inherent problem with our end-to-end detector is the absence of a score for detected boxes that could be used to rank them. Using the likelihood/logits of the initial coordinate tokens is not ideal, as they indicate the distribution over potential starting points rather than an independent probability. Similarly, class-token logits only provide a distribution over class choices, not the probability of the box's existence. Considering text tokens is also impractical, as they represent the actual text rather than the existence of the surrounding box. Therefore, our predictor does not generate a box score. On one hand, as there is no over-prediction, no subsequent filtering or post-processing (such as non-maximum suppression) as well as no score is necessary. On the other hand, this makes comparison on the average precision metric challenging, as when considering all of the predicted bounding boxes jointly, only a single recall level exists for \eclair, making area calculation not meaningful. 

\begin{figure}[h]
    \centering
    \includegraphics[width=0.45\textwidth]{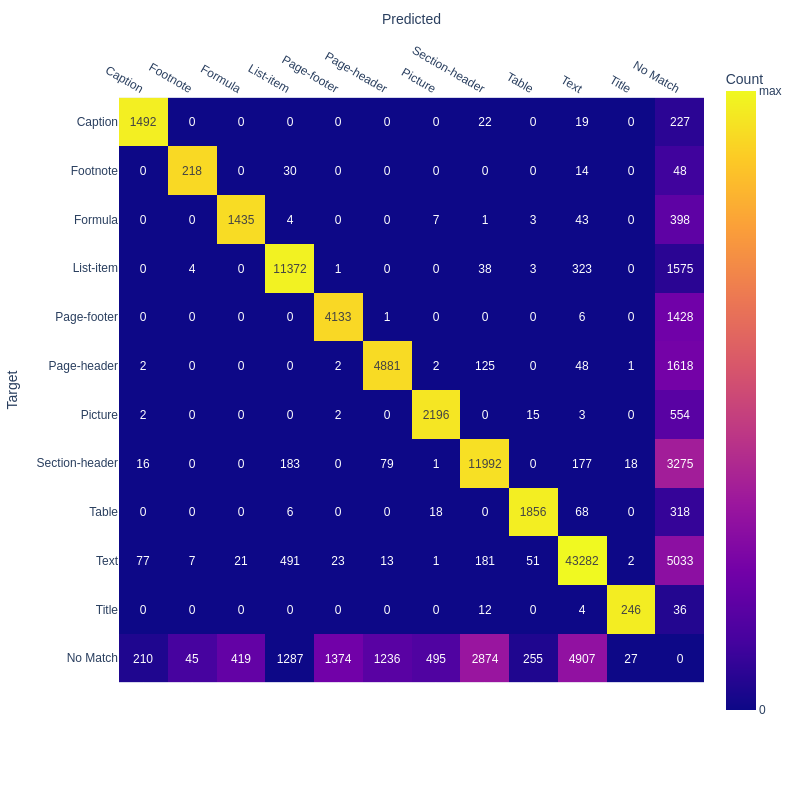}
    \caption{Confusion matrix for \eclair boxes matched with ground truth on the DocLayNet evaluation dataset, averaged over thresholds of $IoU \geq \{0.5, 0.55, ..., 0.9, 0.95\}$.}
    \label{fig:eclair_conf_mat}
\end{figure}

Therefore, it can be seen that comparing \eclair against other works on the $mAP$ metric poses challenges:

\begin{enumerate}
    \item AP is the area under the PR-Curve, which degenerates to a single point without the possibility to rank predictions, making the calculation of the area not meaningful.
    \item The COCO implementation assumes scores are unique. Identical scores (as in our case) lead to incorrect PR-Curves and inconsistent results. \footnote{See \url{https://github.com/MiXaiLL76/faster_coco_eval/issues/46} and \url{https://github.com/cocodataset/cocoapi/issues/678}}.
    \item COCO mAP is computed per class independently (i.e. first separate classes, then match boxes). We propose to first match boxes over all classes and then compute the per-class precision/recall, which allows us to plot a confusion matrix, to better visualize problematic cases. Precision/recall and confusion matrix can still be averaged over multiple IoUs (0.5 to 0.95) like the COCO framework does. For an example of this on the DocLayNet evaluation dataset, see \cref{fig:eclair_conf_mat}.
\end{enumerate}

\begin{figure}[h]
    \centering
    \includegraphics[width=0.45\textwidth]{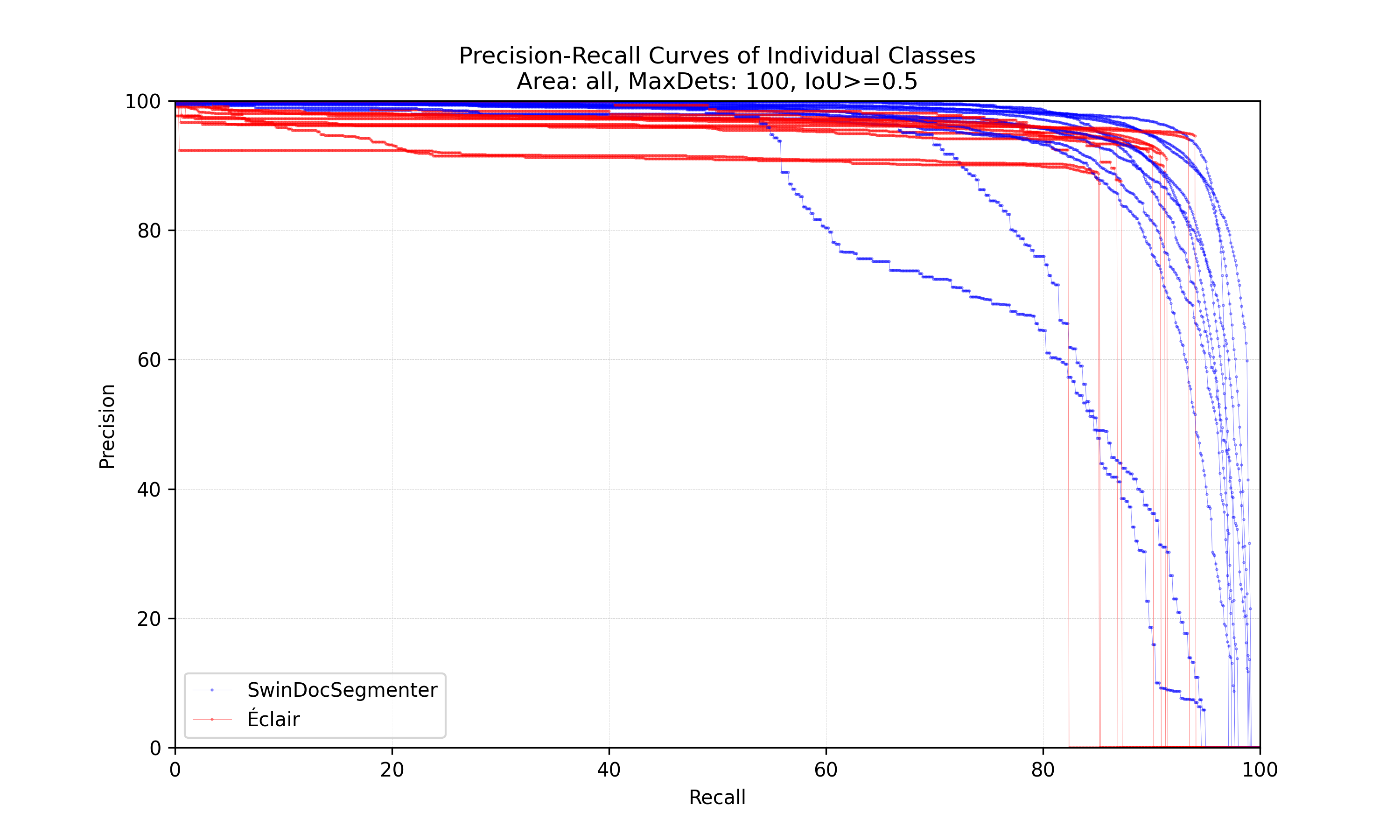}
    \includegraphics[width=0.45\textwidth]{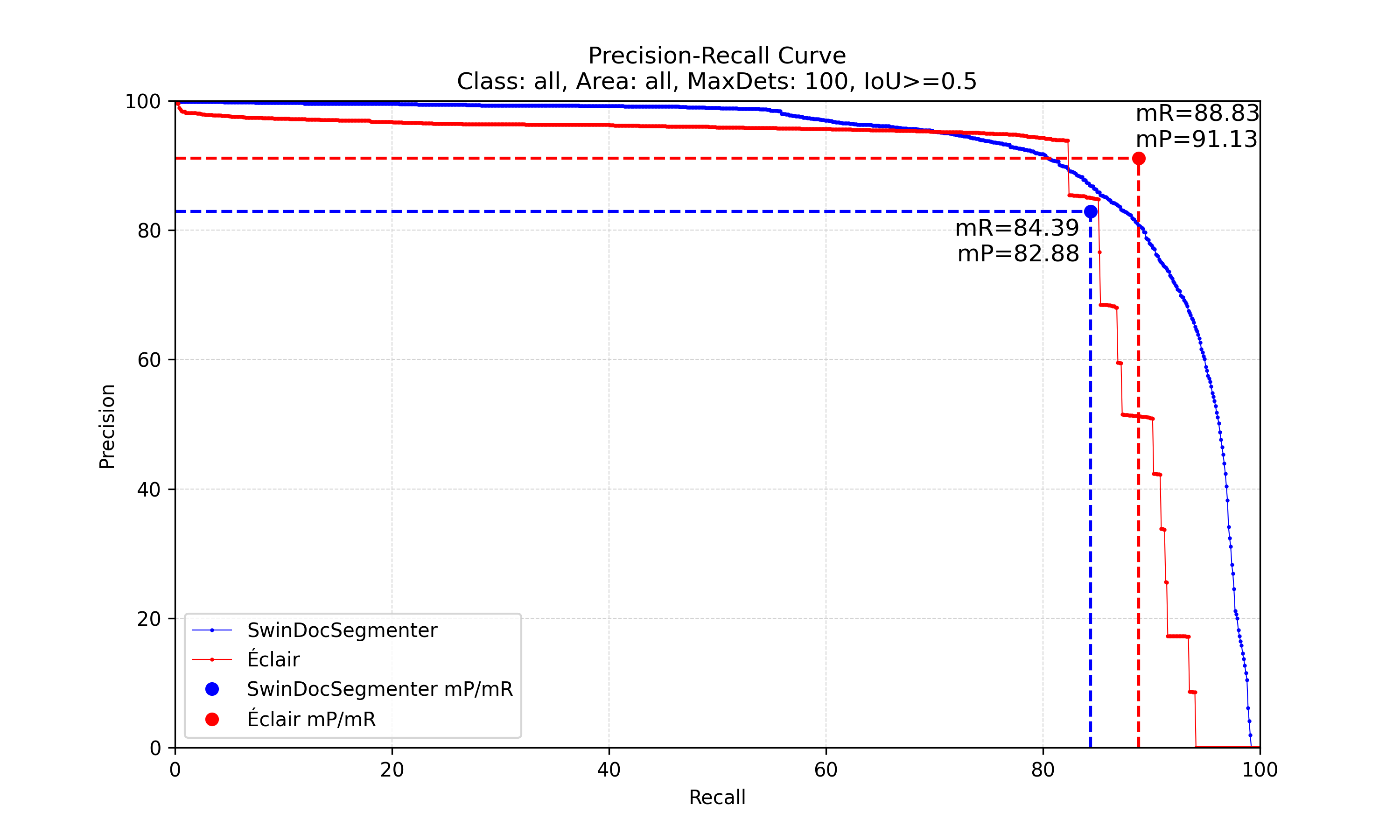}
    \caption{PR-Curves for individual classes and averaged over all classes for $IoU \geq 0.5$ and 1001 recall-bins evaluated on the DocLayNet evaluation dataset. For \eclair, the scores are taken from the class-token logits. The mean precision and recall are taken from \cref{tab:eclair_vs_swindoc_tab}.}
    \label{fig:eclair_vs_swindoc_plot}
\end{figure}

\Cref{fig:eclair_vs_swindoc_plot} shows PR-Curves of individual classes for $IoU \geq 0.5$ and 1001 recall-bins vs predictions from SwinDocSegmenter \cite{swindocsegmenter}. For \eclair, scores are taken from class-token logits, which apparently are not a good separator for true positives vs false positives, compared to SwinDocSegmenter's scores. \eclair does not predict additional low-score boxes, causing curves to drop to zero when all boxes are included. SwinDocSegmenter curves are smoother due to over-predicted boxes, allowing proper score thresholds for each class.

The second part of \cref{fig:eclair_vs_swindoc_plot} shows averaged PR-Curve over all classes for $IoU \geq 0.5$ and 1001 recall-bins. Steps in the \eclair curve come from averaging over classes, cutting off at the mean precision (mP)/mean recall (mR) point. 

In the main paper, we have shown \eclair to be competitive on mAP metric when using methods from the earlier autoregressive object detection literature \cite{pix2seq} such as sequence augmentation and top-k selection that are able to increase the recall of the model at the cost of the low precision. Still, we show \eclair to be a strong predictior without the necessity to introduce such augmentations in the next subsection.

\subsection{Comparisons at corresponding precision/recall levels}
We additionally present a comparison of \eclair to SwinDocSegmenter in a point-to-point manner. Given \eclair's precision and recall value of each class when considering all the predicted bounding boxes, we compare the recall and precision of SwinDocSegmenter at corresponding precision and recall levels (that is, considering equal recall, we want to evaluate which model achieves higher precision, and vice versa). We evaluate each class separately, and report the mean precision and mean recall for both methods. Here, we train a standard \eclair model without sequence augmentation and perform no filtering or post-processing on the predicted bounding boxes, reporting the mean precision and mean recall of the full prediction set. The results are summarized in \cref{tab:eclair_vs_swindoc_tab}. As can be seen, \eclair achieves higher precision at the same recall as SwinDocSegmenter, as well as higher recall at the same precision point, for the vast majority of the classes, as well as on average. 

\begin{table}[h]
    \centering
    \footnotesize
    \begin{tabular}{l|cc|cc}
\multicolumn{5}{c}{$IoU\geq 0.5$} \\
\multirow{2}{*}{Class} & \multicolumn{2}{c|}{\textbf{mP}} & \multicolumn{2}{c}{\textbf{mR}} \\
                & \eclair & SDS* & \eclair & SDS* \\\hline
Caption         & \textbf{89.2} & 83.2 & \textbf{91.2} & 89.1 \\
Footnote        & \textbf{92.4} & 65.6 & \textbf{82.1} & 70.8 \\
Formula         & \textbf{91.7} & 78.7 & \textbf{90.9} & 83.6 \\
List-item       & \textbf{91.0} & 87.8 & \textbf{91.4} & 89.6 \\
Page-footer     & \textbf{94.6} & 93.3 & \textbf{94.0} & 93.2 \\
Page-header     & 93.4 & \textbf{96.7} & 86.8 & \textbf{91.1} \\
Picture         & 86.9 & \textbf{92.8} & 85.2 & \textbf{90.9} \\
Section-header  & \textbf{93.0} & 90.5 & \textbf{90.1} & 87.8 \\
Table           & \textbf{88.4} & 87.8 & \textbf{85.2} & 84.8 \\
Text            & \textbf{94.0} & 91.0 & \textbf{93.4} & 90.9 \\
Title           & \textbf{87.8} & 44.2 & \textbf{87.0} & 56.6 \\\hline
All             & \textbf{91.1} & 82.9 & \textbf{88.8} & 84.4 \\
\end{tabular}
\begin{tabular}{l|cc|cc}
\multicolumn{5}{c}{\vspace{\baselineskip}} \\
\multicolumn{5}{c}{$IoU\geq 0.5:0.95$} \\
\multirow{2}{*}{Class} & \multicolumn{2}{c|}{\textbf{mP}} & \multicolumn{2}{c}{\textbf{mR}} \\
                & \eclair & SDS* & \eclair & SDS* \\\hline
Caption         & \textbf{82.8} & 72.1 & \textbf{84.7} & 79.5 \\
Footnote        & \textbf{79.0} & 59.8 & \textbf{70.1} & 62.8 \\
Formula         & \textbf{76.8} & 56.5 & \textbf{75.8} & 56.1 \\
List-item       & \textbf{85.0} & 68.2 & \textbf{85.4} & 82.1 \\
Page-footer     & \textbf{74.7} & 45.3 & \textbf{74.2} & 59.8 \\
Page-header     & \textbf{78.6} & 64.0 & \textbf{73.0} & 68.4 \\
Picture         & 80.7 & \textbf{88.6} & 79.1 & \textbf{85.3} \\
Section-header  & \textbf{78.7} & 53.4 & \textbf{76.2} & 59.8 \\
Table           & \textbf{84.9} & 84.3 & \textbf{81.8} & 81.5 \\
Text            & \textbf{88.5} & 60.6 & \textbf{88.0} & 79.6 \\
Title           & \textbf{83.2} & 20.6 & \textbf{82.3} & 49.7 \\\hline
All             & \textbf{81.2} & 61.2 & \textbf{79.2} & 69.5 \\
    \end{tabular}
    \caption{The mean precision and mean recall of \eclair for each class and the corresponding mean recall and mean precision of SwinDocSegmenter for the respective recall/precision on the PR-curve evaluated on the DocLayNet evaluation dataset. Computed for $IoU \geq 0.5$ (corresponding to \cref{fig:eclair_vs_swindoc_plot}) and for averaged thresholds of $IoU \geq \{0.5, 0.55, ..., 0.9, 0.95\}$ (default for COCO metrics). *SDS: SwinDocSegmenter.}
    \label{tab:eclair_vs_swindoc_tab}
\end{table}

\section{LLM Training}

We train both models with a total of 300B tokens obtained from a combination of sources. We ensure that the models are trained for 3.3 epochs of the tokens  extracted using \eclair and PyMuPDF4LLM. These tokens are extracted from a common set of PDFs for both methods. The rest of the 300B training tokens come from various sources including CommonCrawl snapshots, Stack Exchange, OpenWebMath~\cite{paster2023openwebmath}, PubMed Abstracts, PubMed Central, bioRxiv, SEC filings, Wikipedia and ArXiv data. 

\subsection{Postprocessing and Joining Pages}
To join pages and handle the positioning of floating objects, we follow these steps:

\begin{enumerate}
    \item \textbf{Process Pages Individually:} Each page is processed separately. To manage paragraphs that span across pages, we need to carry open paragraphs over to next pages.
    \item \textbf{Reassign Floating Objects:} Floating objects (e.g., images, tables, captions) are removed and captions are reassigned to their respective objects using Hungarian matching based on the Manhattan distance of the bounding boxes.
    \item \textbf{Concatenate Pages:} Pages are concatenated while skipping sections like Table of Contents, Bibliography, and Indexes by detecting typical headings. Floating text blocks (e.g., Text and List-item) are merged based on specific rules, such as not ending with punctuation.
    \item \textbf{Remove Markdown Formatting:} All markdown formatting is removed from the inner text to ensure consistency.
    \item \textbf{Flush Floating Objects:} After processing each page, floating objects that are not part of the floating text are flushed to the output blocks.
\end{enumerate}

\section{Examples of predictions}

\begin{figure*}[htbp]
    \centering
    \begin{minipage}[c]{0.33\linewidth}
        \centering
        \includegraphics[width=\linewidth]{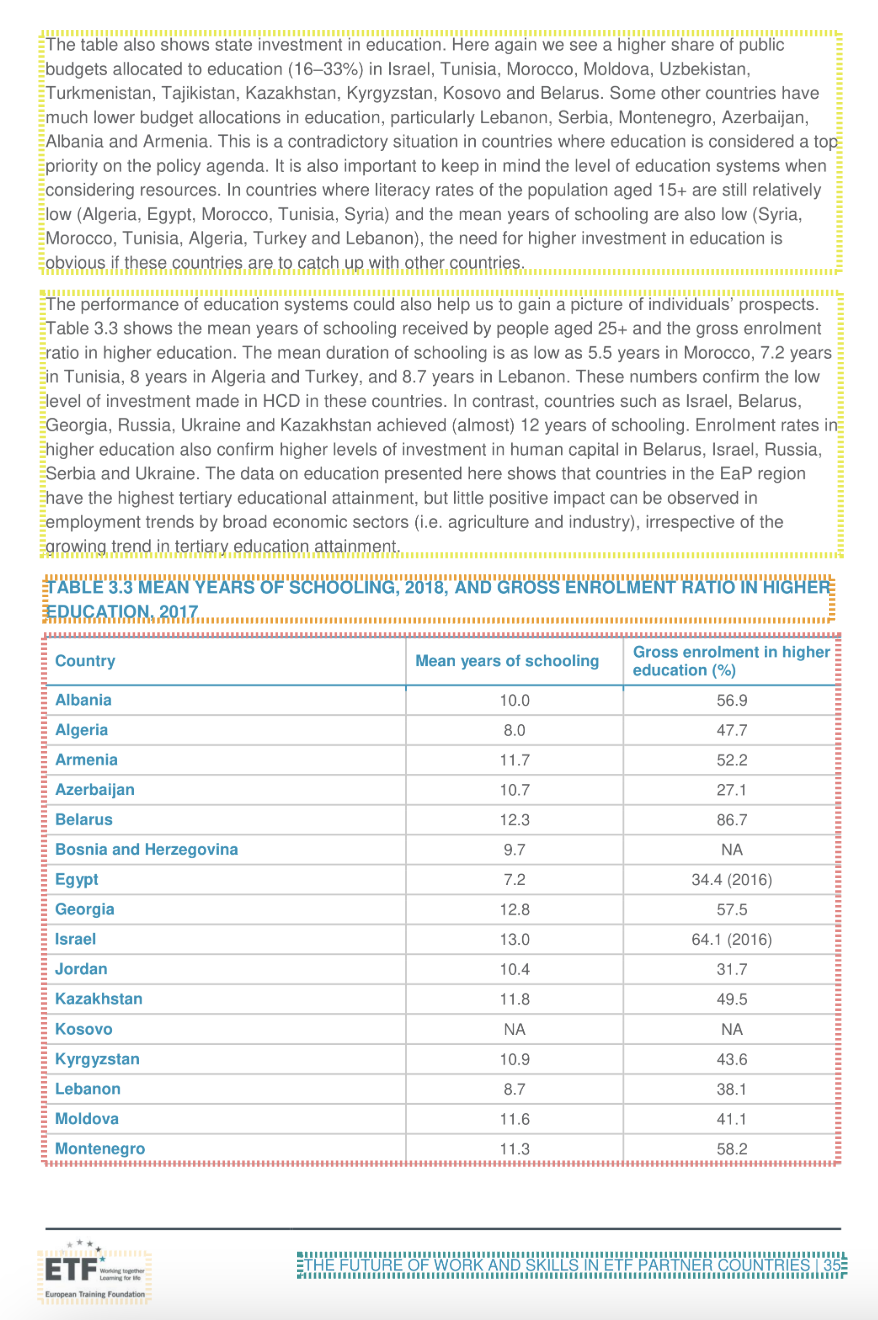}
    \end{minipage} \hspace{0.5cm}
    \begin{minipage}[c]{0.57\linewidth}
        \centering
        \includegraphics[width=\linewidth]{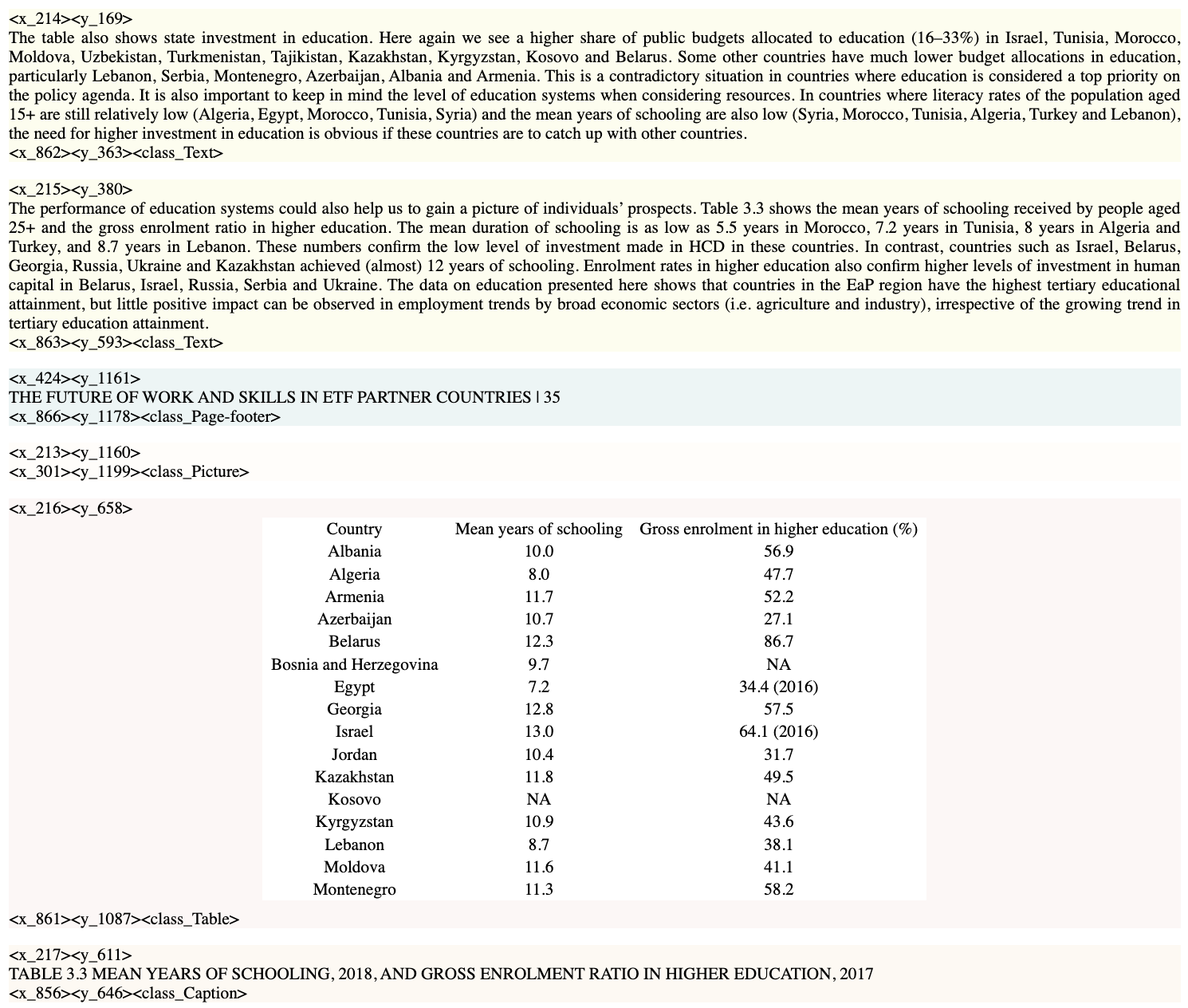}
    \end{minipage} \\[0.5cm]  
    
    \begin{minipage}[c]{0.34\linewidth}
        \centering
        \includegraphics[width=\linewidth]{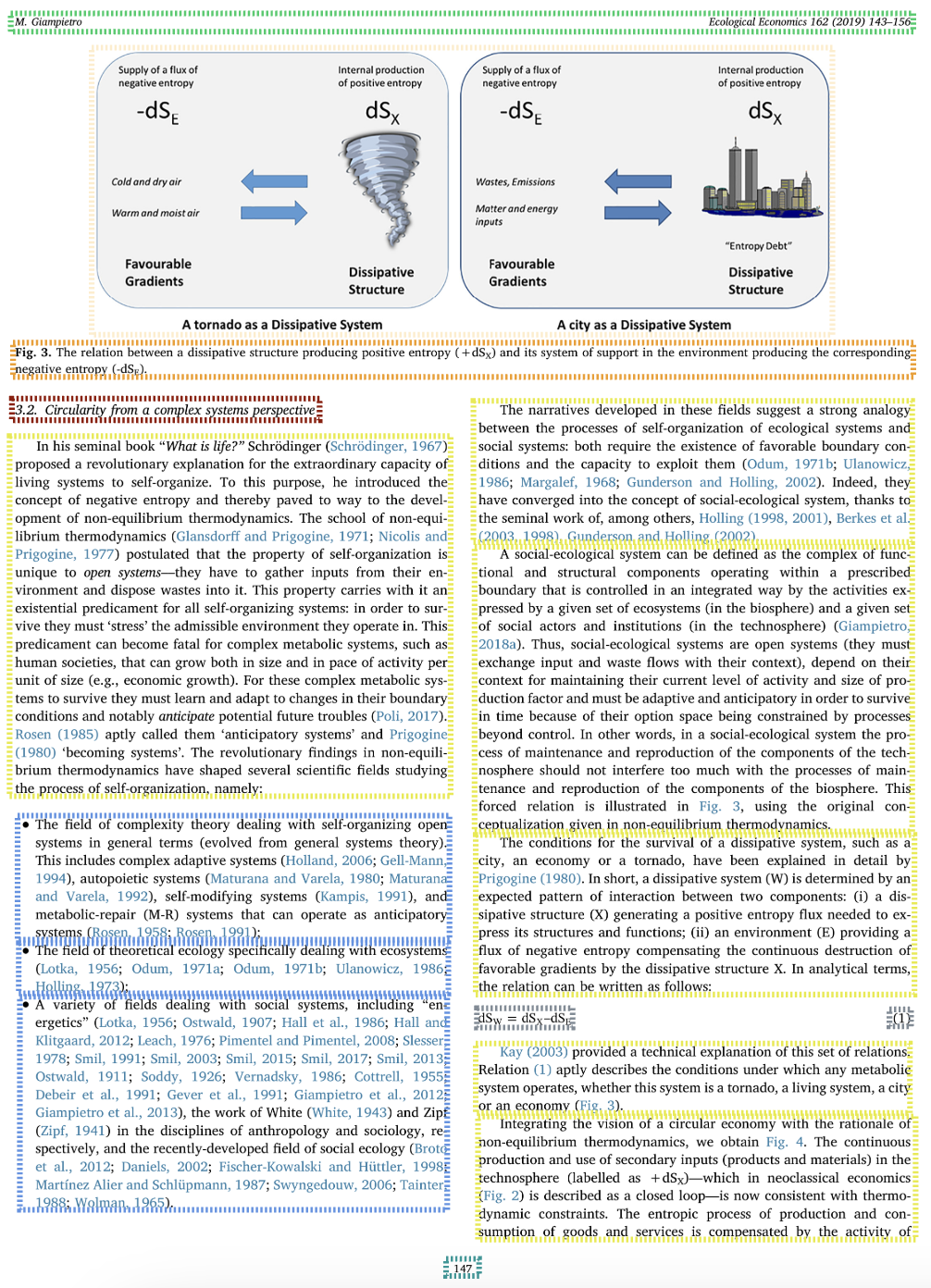}
    \end{minipage} \hspace{0.5cm}
    \begin{minipage}[c]{0.57\linewidth}
        \centering
        \includegraphics[width=\linewidth]{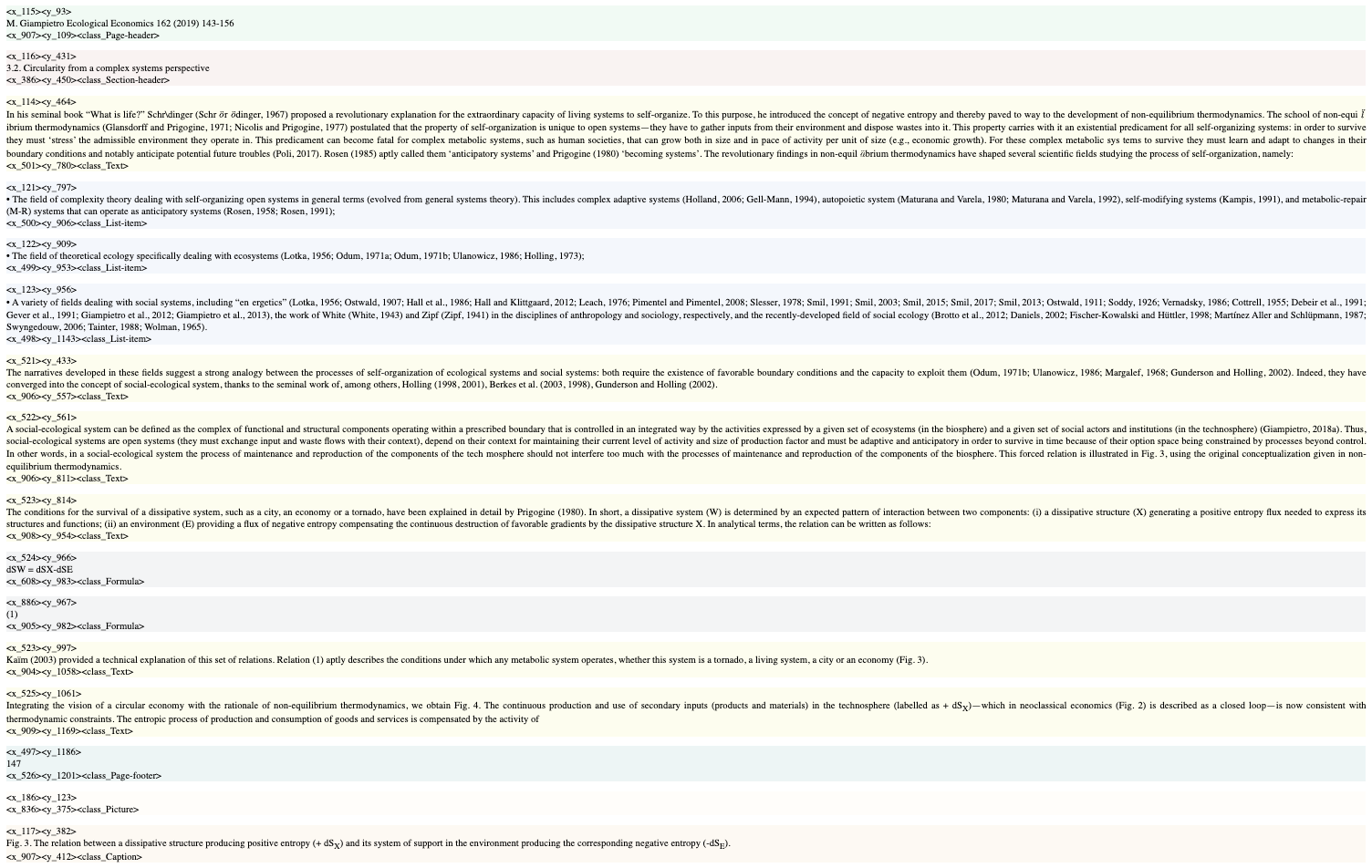}
    \end{minipage}
    \caption{Examples of pages with tables, formulae and pictures. On the left, predicted bounding boxes superimposed on the original sample image. On the right, the corresponding full predictions.}
    \label{fig:good-samples}
\end{figure*}

In this section, we present examples of predictions from \eclair on samples from the Common Crawl dataset. Figure~\ref{fig:good-samples} contains samples with tables, formulae, pictures and a variety of other elements. 

\end{document}


%
%
%
sa
\todo[inline]{Add some information about reading order heuristics for DocLayNet, table autolabeling for plain datasets and steps to obtain the GitHUb Readme dataset in supplementary material}

\section{Examples of predictions}

\IK{shall we show 4-5 almost full-page predictions from Lukas' visualisation tool?}

\section{Training Details}

\subsection{Hallucinations}

It is a common observation \cite{something} that generative models may hallucinate content, and there have been various attempts to reduce the likelihood of hallucinations by introducing grounding \cite{something} methods. Some document-level OCR models such as Nougat \cite{Nougat} detect hallucinations by tracking a moving average of logits and flagging the outputs when a certain threshold is reached. However, such an approach may not be suitable if the model predicts hallucinations with high certainty. 

\eclair introduces a hallucination mitigation strategy characterized using three components: (i) the inference-time prompt is always the Maximal Informative Prompt (MIP), enforcing a strict syntax which would reject non-compliant boxes, (ii) the syntax is semantically grounded to the visual document structure rather than being a generic delimiter such as \verb|<bbox></bbox>| tags, \KC{I am not sure I understand the second point...}(iii) the spatial and categorical validity is enforced by verifying that the bottom-right corner of each bounding box exceeds the top-left corner and that classes conform to a validated schema. This approach allows \eclair to effectively filter out invalid boxes while preserving accurate ones, thereby ensuring comprehensive and reliable page predictions. By implementing this layered filtering strategy, we achieve substantial reduction in model hallucinations.

\PF{Here we say how we can filter boxes, but we are not saying how much it helps. I think it's crucial to say that we can prevent a lot of hallucinations and give some evidence.
\KS {I updated paragraph above but giving evidence of statistics might be hard, considering its property of layout and foreign language and then you need to define how much of page it extracted ? wdyt}} \KC{maybe can say: While quantifying the effect of hallucinations is challenging, by implementing this layered filtering strategy, we qualitatively observe a substantial reduction in model hallucinations.} 

\section{pre-processing steps for the training datasets }
\section{Joining Pages}

\section{mAP - Back to Simple Metrics}
\label{sec:map}

The problem \TR{challenge} with our proposed end-to-end detector is, that we do not provide a score for a detection box. One could think about using the likelihood/logits of the first coordinate tokens, but these are more likely a distribution over the next possible starting points, even if calibrated, thus not useful as an independent/unordered probability. The other possibility could be the class token logits, but those will only give a distribution over the choice of classes, not really giving a probability of the box existing at all. The last option could be to consider the text tokens, but then again, these are more a representation of actual text being read, and not representing the existence of the box around. Consequently, for our predictor, there is no box score. Still, for comparing against other works, which use the mAP for this comparison, it is problematic: The first problem lies in the nature of (m)AP: AP is the area under the PR curve, which is degenerated for the case where there is no score (or using score=1 for all predictions), it'll just be a single point (thus can be reduced to \(\text{AP} = \text{precision} (@\text{recall}=1)\)). The second issue lies in the usage of the COCO implementation, which sorts the predictions by score and based on that computes the PR curve, assuming that scores are never exactly identical. If all scores are the same (which is the case for our end-to-end detector), the computed curve will depend on the order of predictions, as coco will traverse on the PR curve prediction-by-prediction, thus it does not produce the correct PR curve and will even give different results for different ordering \footnote{See also submitted bug report \url{https://github.com/MiXaiLL76/faster_coco_eval/issues/46} and \url{https://github.com/cocodataset/cocoapi/issues/678}}. Next, the nature of the PR-Curve is meant to be able to set a score, such that predictions can be filtered to achieve the precision/recall represented on that curve. Since we predict boxes as part of the output token stream, we would not want to discard boxes in general, so we do not have a value which would discard a box. Lastly, the COCO mAP is computed per class, i.e. each class is considered independently. Instead, we find it more useful to first match the boxes and then compute the class correctness. This effectively makes the matching harder, because competing boxes are considered across classes instead of within a class. On the other hand, this does provides more detailed information on the error cases if plotting a confusion matrix. Thus, we propose computing the mean precision/recall directly and also plotting a confusion matrix (based on a fixed score for other work), better visualizing the problematic cases. The precision/recall and confusion matrix may still be averaged over multiple IoUs (between 0.5 and 0.95) like the COCO framework in order to factor out the threshold. An implementation is provided in the appendix.


\begin{figure*}[t]
    \centering
    \includegraphics[width=0.45\textwidth]{images/PR_Curve_catall_iou0.5to0.95_areaall_maxDet100.pdf}
    \caption{{\todo[inline]{Update the graphics and reduce space}} Precision recall graph based on COCO-mAP computations and smoothing, with 1000 steps (as opposed to the default 100 steps) for recall, showing the respective levels for the precision and recall of \eclair. Computing the COCO-mAP for SwinDocSegmenter, we get $\text{mAP}=75.4$, $\text{mAR}=84.1$; It can be argued, that \eclair is surpassing SwinDocSegmenter by a large margin. \KC{potentially can just report numbers in a table and move curves to supplementary}}
    \label{fig:eclair_doclaynet_pr_fig}
\end{figure*}

\section{LLM Training}

\subsection{misc}

talk about multi-token inference measurement. 

Following the observation in \cite{hu2024mplugdocowl15unifiedstructure} that text is more correlated within a line rather than across lines, we compress the sequence with a horizontal convolutional kernel of size $2\times 8$ and stride $8$. Given an input image of size $1280\times 1024$ which produces $80\times 64$ patches, the sequence length is reduced from $5120$ to $1280$

We used seq length 3.4K